\pdfoutput=1
\documentclass[11pt]{article}

\usepackage[final]{acl}

\usepackage{times}
\usepackage{latexsym}

\usepackage[T1]{fontenc}
\usepackage[utf8]{inputenc}
\usepackage{enumitem}
\usepackage{microtype}
\usepackage{times}
\usepackage{soul}
\usepackage{url}
\usepackage{times}
\usepackage{soul}
\usepackage{url}
\usepackage{caption}
\usepackage{tikz}
\usepackage{svg}
\usepackage{amssymb}
\usepackage{adjustbox}
\usepackage{subcaption}
\usepackage{graphicx}
\usepackage{inconsolata}
\usepackage{svg}
\usepackage{booktabs, multirow}
\usepackage[most]{tcolorbox}
\title{M3Hop-CoT: \textit{M}isogynous Meme Identification with \textit{M}ultimodal \textit{M}ulti-hop \textit{C}hain-of-\textit{T}hought}

\author{Gitanjali Kumari$^1$ \quad Kirtan Jain $^1$\quad Asif Ekbal$^2$\\
$^1$Department of Computer Science and Engineering,\\ $^1$Indian Institute of Technology Patna, India\\
$^2$School of AI and Data Science, Indian Institute of Technology Jodhpur, India\\
{\tt \small $^1\{$gitanjali\_2021cs03,kirtan\_2101cs38$\}$@iitp.ac.in,$^2$asif@iitj.ac.in} \\
}

\begin{document}
\maketitle
\begin{abstract}
In recent years, there has been a significant rise in the phenomenon of hate against women on social media platforms, particularly through the use of misogynous memes. These memes often target women with subtle and obscure cues, making their detection a challenging task for automated systems. Recently, Large Language Models (LLMs) have shown promising results in reasoning using Chain-of-Thought (CoT) prompting to generate the intermediate reasoning chains as the rationale to facilitate multimodal tasks, but often neglect cultural diversity and key aspects like emotion and contextual knowledge hidden in the visual modalities. To address this gap, we introduce a \textbf{M}ultimodal \textbf{M}ulti-hop CoT (M3Hop-CoT) framework for \textbf{M}isogynous meme identification, combining a CLIP-based classifier and a multimodal CoT module with entity-object-relationship integration. M3Hop-CoT employs a three-step multimodal prompting principle to induce emotions, target awareness, and contextual knowledge for meme analysis. 
Our empirical evaluation, including both qualitative and quantitative analysis, validates the efficacy of the M3Hop-CoT framework on the SemEval-2022 Task 5 (\textbf{MAMI task}) dataset, highlighting its strong performance in the macro-F1 score.
Furthermore, we evaluate the model's generalizability by evaluating it on various benchmark meme datasets, offering a thorough insight into the effectiveness of our approach across different datasets \footnote{Codes are available at this link: \url{https://github.com/Gitanjali1801/LLM_CoT}}.
\end{abstract}
\section{Introduction}
In recent years, the proliferation of memes on social media platforms like Facebook, Twitter, and Instagram has gained significant attention due to their widespread influence and potential to shape public discourse. While many memes are created for entertainment, some serve political or activist purposes, often employing dark humor. Misogynous memes\footnote{\textcolor{red}{WARNING: This paper contains meme samples with slur words and sensitive images.}}, however, stand apart by propagating hatred against women through sexist and aggressive messages on social media \cite{attanasio-etal-2022-milanlp, zhou-etal-2022-dd-tig,arango-etal-2022-hateu}.
% zhang-wang-2022-srcb,zhou-etal-2022-dd-tig,chen-chou-2022-rit,fersini-etal-2022-semeval}.
These memes exacerbate sexual stereotyping and gender inequality, mirroring offline societal issues \cite{Franks} and have become a concerning issue \cite{chen-chou-2022-rit,zhang-wang-2022-srcb,fersini-etal-2022-semeval}.\\
\begin{figure}[t!]
\centering
\includegraphics[width=\linewidth]{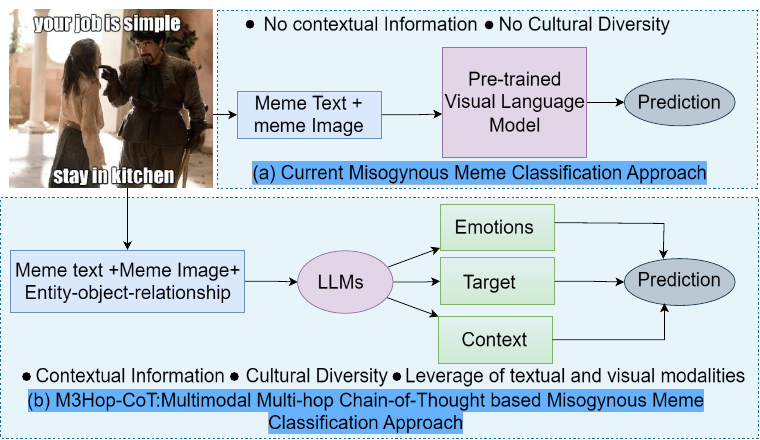}
    \caption{Comparison between (a) fine-tuning visual language model approach and (b) Chain-of-Thought based approach.}\label{fig:intro_exam_analysis}
\end{figure}
\noindent
Identifying misogynous memes is much more challenging than other memes, as the task demands an understanding of world knowledge and common sense \cite{10.1145/3656169}. Despite the challenges, developing deep learning models to classify such memes can provide sociological benefits, such as understanding hidden meanings, supporting humanities research, and raising awareness on a large scale \cite{kumari-etal-2024-eacl}. Previous research has primarily focused on developing robust deep-learning models that learn cross-modal interactions (c.f. Figure \ref{fig:intro_exam_analysis} (a)) from scratch to identify these memes \cite{ rijhwani-etal-2017-estimating,sharma-etal-2020-semeval,NEURIPS2020_1b84c4ce, suryawanshi-etal-2020-multimodal,pramanick-etal-2021-detecting, hossain-etal-2022-mute,sharma-etal-2022-disarm}. However, learning complex multimodal interactions can be difficult with limited data \cite{10.1145/3581783.3612498}. The advent of Large Language Models (LLMs) offers a way to bridge this gap. Although LLMs are highly adept at question-answering and reasoning tasks, they often overlook the cultural diversity of human reasoners, crucial for tasks demanding commonsense, contextual knowledge, and multimodal reasoning \cite{li-zhang-2023-cultural,10.1162/tacl_a_00634}. However, the recent concept of \textbf{Chain-of-Thought (CoT)} has demonstrated the potential of LLMs for multi-hop reasoning \cite{wei2023chainofthought,fei-etal-2023-reasoning,wu-etal-2023-chain}, showing that LLMs can perform chain-style reasoning effectively with the right prompts. Nonetheless, most CoT reasoning studies focus primarily on the language modality \cite{lu2023dynamic, DBLP:journals/corr/abs-2207-00747}, often overlooking multimodal contexts. Analyzing memes is particularly challenging because their implicit meanings are not fully conveyed through text and images. In such a scenario, neglecting one modality in meme detection can negatively impact model performance.\\
\noindent
As depicted in Figure \ref{fig:intro_exam_analysis} (b), if an LLM can not only interpret emotions, such as anger or disgust from the text, ``Your job is simple. Stay in Kitchen," but also analyze the visual elements of the meme featuring a woman and a man, which would further enhance its ability to recognize emotions by considering their facial expressions and body language. This is crucial for identifying sexist stereotypes. Moreover, the LLM can determine if the meme targets women by evaluating both textual and visual modalities. Furthermore, understanding the broader context, which encompasses societal and cultural discussions on gender roles and equality, is crucial, and this can also be achieved using LLMs. To achieve this, the ability to perform \textbf{multi-hop reasoning Chain-of-Though(CoT)} (i.e., inferring emotion, target, and then understanding the context) is indispensable. By hierarchically considering key aspects of misogynous memes, such as \textit{emotions}, \textit{targets}, and \textit{contextual backgrounds}, we can create general-purpose models that are in sync with human intent for real-world tasks like meme identification.\\
\noindent
Our proposed work is motivated by the aforementioned discussion, where we introduce a deep learning-based framework named M3Hop-CoT (\textbf{M}isogynous Meme Identification with \textbf{M}ultimodal \textbf{M}ulti-\textbf{hop} \textbf{C}hain-of-\textbf{T}hought) a modular approach that leverages an LLM as the “reasoning module” and operates over a given meme. In the M3Hop-CoT approach, we first extract the Entity-object-relationship (EORs) of the meme-image using a scene graph \cite{DBLP:journals/corr/abs-2002-11949}. Subsequently, the meme text, image, and EORs are fed into the multi-hop CoT LLM, enabling it to identify three crucial hidden cues for inferring the meme's rationales: (i) emotion, (ii) target, and (iii) context. 
M3Hop-CoT eliminates the need for external resources, also bridging the gap between the modalities by utilizing both textual and visual aspects of the meme in rational generation at zero cost. 
To ensure the weighted contribution of each reasoning step, we employ a hierarchical cross-attention mechanism that assesses the contribution of each rationale in decision-making. \\
% On top of it, M3Hop-CoT enhances multimodal rationale representations while training by jointly incorporating supervised contrastive learning (SCL) and cross-entropy loss. SCL helps bring instances of the same class closer in semantic space by concentrating on their latent representation.\\
\noindent
The main contributions of this work are summarized below: (i) This is the ﬁrst study where we introduce multimodal LLM in a CoT manner to identify the misogynous memes. (ii) We introduce the M3Hop (Misogynous Meme Identification with Multimodal Multi-hop Chain-of-Thought) framework, where we utilize the meme text and EORs of the meme-image as a prompt to the LLM in a multi-hop CoT manner, enabling it to identify three crucial rationales helpful to detect misogynous memes: (a) emotion, (b) target, and (c) context. (iii) Our empirical evaluation, including both qualitative and quantitative analysis, validates the efficacy of the M3Hop-CoT framework on several datasets, highlighting its strong performance. 
\section{Related Work}
\textbf{Detection of Misogynous memes. }
Previous studies on memes have predominantly focused on identifying hate or offensive content \cite{rijhwani-etal-2017-estimating,sharma-etal-2020-semeval,NEURIPS2020_1b84c4ce,suryawanshi-etal-2020-multimodal,sharma-etal-2022-disarm,hossain-etal-2022-mute, yadav-etal-2023-towards}. While most of the existing meme research has focused on refining multimodal representations by exploring interactions between textual and visual elements \cite{kumari-etal-2021-co,Akhtar2022AllinOneES,sharma-etal-2022-domain,10.1007/978-3-031-28244-7_7,sharma-etal-2023-memex}, still error analyses in these studies have revealed a significant gap in the contextual comprehension of memes \cite{cao-etal-2022-prompting}. While existing research on detecting misogynous content has largely focused on unimodal data (primarily text) \cite{10.1145/2908131.2908183,Fersini2018OverviewOT,10.1145/3350546.3352512}, the integration of multimodality (text and image), on the other hand, is still a work in progress \cite{zhou-etal-2022-dd-tig, zhi-etal-2022-paic, arango-etal-2022-hateu,singh-etal-2023-female}. \\ %chen-chou-2022-rit, rao-rao-2022-asrtrans, zhang-wang-2022-srcb,srivastava-2022-poirot-semeval,
\noindent
\textbf{Large Language Models. }Pre-training of language models has garnered significant interest for its ability to enhance downstream applications \cite{DBLP:journals/corr/abs-1910-10683}. Recently, large-scale language models (LLMs), such as GPT-3 \cite{kojima2023large}, ChatGPT \cite{ouyang2022training}, LLaMA \cite{touvron2023llama} etc., have demonstrated remarkable potential for achieving human-like intelligence.
LLMs have shown exceptional capabilities in common-sense understanding \cite{paranjape-etal-2021-prompting,liu-etal-2022-generated}  with the incorporation of the chain of thought (CoT) method which has revolutionized the way machines approach reasoning-intensive tasks \cite{wei2023chainofthought,zhou2023leasttomost}. \\
\noindent
% While previous research has often struggled with incorporating external knowledge and commonsense understanding in a meme classifier and has been limited by fitting spurious correlations between multimodal features, our proposed model \textbf{M3Hop-CoT} bridges this gap with a novel prompt-based approach. Our method distinguishes itself from the existing techniques by employing a \textit{Multimodal Multihop CoT} based approach to simultaneously analyze the meme text and the entity-object relationship of the meme image, thereby deciphering the emotional, targeted, and contextual dimensions of a misogynous meme. By doing so, we aim to integrate culturally diverse reasoning into our proposed misogynous meme classifier.
While previous research has often struggled with incorporating external knowledge and commonsense understanding and has been limited by fitting spurious correlations between multimodal features, our proposed model \textbf{M3Hop-CoT} bridges this gap with a novel prompt-based approach by employing a \textit{Multimodal Multihop CoT} based approach to simultaneously analyze the meme text and the entity-object relationship of the meme image, thereby deciphering the emotional, targeted, and contextual dimensions of a misogynous meme. By doing so, we aim to integrate culturally diverse reasoning into our proposed misogynous meme classifier.
\section{Dataset}
For our experiments, we employ two misogynous meme datasets: MAMI (SemEval2022 Task 5, Subtask A) (in English) \cite{fersini-etal-2022-semeval} and MIMIC (in Hindi-English Code-Mixed) \cite{10.1145/3656169} (Refer to Table \ref{tab:stat}). To demonstrate the generalizability of our CoT-based approach, we conduct experiments on three benchmark meme datasets: Hateful Memes \cite{DBLP:journals/corr/abs-2005-04790}, Memotion2 \cite{memotion2}, and Harmful Memes \cite{sharma-etal-2022-disarm} (See Appendix Table \ref{tab:classwise_dis} for data statistics) 
\begin{table}[ht!]\centering
\scriptsize
\adjustbox{width=\linewidth}{\begin{tabular}{lrrrr}\toprule
\textbf{Dataset} &\textbf{Train set} &\textbf{Test set} & \textbf{Task}\\\midrule
MAMI &10,000 &1,000 &Misogynous Meme Detection \\
MIMIC & 4,044&1,010 &Misogynous Meme Detection \\
Hateful Meme &8,500 &1,000 &Hateful Meme Detection \\
Memotion2 &7,500 &1,500 &Offensive Meme Detection \\
Harmful Meme &3,013 &354 &Harmful Meme Detection \\
\bottomrule
\end{tabular}}\caption{ Dataset Statistics}\label{tab:stat}
% \vspace{-0.7cm}
\end{table}
\section{Methodology}
This section illustrates our proposed M3Hop-CoT model to identify the misogynous meme. The overall workflow of our proposed \emph{M3Hop-CoT} model is shown in Figure \ref{fig:proposed_model}, and its components are discussed below.
\subsection{Problem Formulation}
Let $\mathcal{D} = {(x_i, y_i)}_{i=1}^N$ represent the dataset of misogynous memes, where $N$ is the number of samples, $x_i \in \mathcal{X}$ is the $i$-th meme (comprising text and images), and $y_i \in \{0, 1\}$ is its corresponding misogyny label ($1$ for misogynous, $0$ for non-misogynous). Our objective is to train a classifier $f_\theta: \mathcal{X} \to \mathcal{Y}$ to predict correct misogynous label $\mathcal{\hat{Y}}$, parameterized by $\theta$, to minimize a loss function $\mathcal{L}(\mathcal{\hat{Y}}|\mathcal{X},\theta )$, defined over the output space $\mathcal{Y}$ and the predicted label $\mathcal{\hat{Y}}$.
\subsection{Encoding of Meme} \label{sec:feature}
A meme sample $\mathcal{X}_i$ comprises of meme text $T_i = (t_{i_1}, t_{i_2},\ldots, t_{i_k})$, which is tokenized into sub-word units and projected into high-dimensional feature vectors, where $k$ is the number of tokens in the meme text, and image $I_i$ with regions $r_i = \{r_{i_1},r_{i_2},\ldots,r_{i_N}\}$; for $r_{i_j} \in R^{N}$, where $N$ is the number of regions. These components are input into a pre-trained CLIP model \cite{pmlr-v139-radford21a}
designed to extract features by understanding text and images at a semantic level. 
\begin{equation}\label{eq:rep}
    ft_i, fv_i = CLIP(t_i,r_i)~;
\end{equation}
where $ft_i \in \mathbb{R}^{d_t}$ and $fv_i \in \mathbb{R}^{d_v}$ are the extracted text and visual features, respectively, with $d_t$ and $d_v$ denoting the dimensions of the text and visual feature spaces. To integrate these features, we use the Multimodal Factorized Bilinear (MFB) pooling technique \cite{kumari-etal-2021-co,10.1007/978-3-031-28244-7_7}. The interaction between textual and visual features was limited in earlier fusion techniques \cite{ZHANG2022103877} (e.g., concatenation, element-wise multiplication, etc.). These methods did not allow for comprehensive interaction between textual and visual features, essential for generating robust and nuanced multimodal features. Bilinear pooling, while effective in capturing detailed associations between textual and visual features through outer products, introduces high computational costs and risks of overfitting due to the large number of parameters required \cite{yu2018beyond}. In contrast with this, MFB provides an efficient solution by factorizing the bilinear pooling operation. This approach effectively maximizes the association between textual and visual features while mitigating the computational and overfitting concerns associated with traditional bilinear pooling \cite{yu2017multi,Kumari2023}. \\
MFB combines $ft_i$ and $fv_i$ to produce a multimodal representation $M_i$ with dimensions $\mathbb{R}^{o \times 1}$. The MFB module employs two weight matrices, $U$ and $V$, to project and sum-pool the textual and visual features, respectively. The resulting fusion is expressed in the following equation:
\begin{equation}\label{sumpool}
\begin{aligned}
M_i = \text{SumPool}(U^Tft_i \circ V^Tfv_i, k)~;
\end{aligned}
\end{equation}
Here, $\circ$ represents element-wise multiplication, and $\text{SumPool}(x,k)$ refers to a sum-pooling operation over $x$ using a one-dimensional non-overlapping window of size $k$.

\begin{figure}[t!]
\centering
\includegraphics[width=\linewidth]{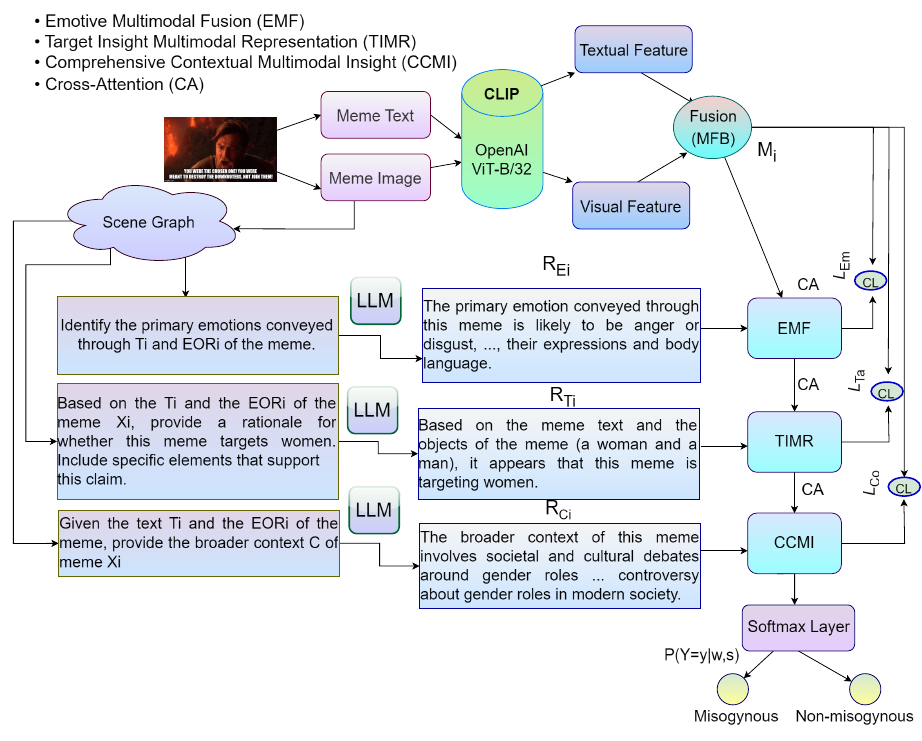}
    \caption{Illustration of the proposed M3Hop-CoT model.}\label{fig:proposed_model}
 % \vspace{-0.7cm}
\end{figure}

\subsection{Entity-Object-Relationships (EoRs) Extraction}
Improving the representation of textual and visual components in meme analysis is crucial to bridge the semantic gap between these modalities. To retrieve a better visual representation, we utilize an unbiased scene graph model proposed by \citet{DBLP:journals/corr/KrishnaZGJHKCKL16}, which leverages Faster RCNN \cite{renNIPS15fasterrcnn} and joint contextual feature embedding to extract unbiased \textit{Entity Object Relationship (EOR)} data from the visual modality of each meme (c.f. Figure \ref{fig:scene_graph_arc}). For a meme image $I_i$ in meme $\mathcal{X}_i$, the scene graph is defined as $T_I \subseteq (E_I \times R_I \times E_I)$, where $T_I$ is the set of visual triples, $E_I$ is the entity set, and $R_I$ is the relation set, with $R_I \subseteq R$. Each entity $e_{I,k} = (e_{t, I,k}, A_{I,k}, b_{I,k}) \in E_I$ consists of the entity type $e_{t, I,k} \in E_t$, where $E_t$ is the set of entity types (Refer to Appendix Figure \ref{fig:scene_graph}). For meme $\mathcal{X}_i$, the extracted entity-object-relation triplets from its scene graph are denoted as $EOR_i$. The notation $\text{EOR}_i$ represents the top $k$ (in this case, k=5) entity-object relations from the scene graph of image $I_i$, expressed as $\text{EOR}_i= (\text{EOR}_i^1,\text{EOR}_i^2,\ldots,\text{EOR}_i^k)$. Integrating this visual understanding into our LLM is intended to uncover such hidden cues in images that are crucial for making informed, human-like decisions for detecting misogynous memes.
\subsection{Chain-of-Thought Prompting}
Our M3Hop-CoT model (refer to Figure \ref{fig:proposed_model}) employs a Chain-of-Thought (CoT) prompting approach \cite{wei2023chainofthought,zhou2023leasttomost} to facilitate multi-step reasoning during meme analysis. Rather than directly querying the LLM for the final label $\hat{y}$, we aim for the LLM to infer detailed information about the meme's emotional content, its potential targeting of women, and the broader context of its creation and interpretation. The three-hop prompts are constructed as follows:\\
\noindent
\textbf{Step 1. }The first prompt queries the LLM about the emotions $\mathcal{E}$ conveyed by the meme with the following template:\\
\noindent
\colorbox{gray!10}{
\begin{minipage}{0.95\columnwidth}
\color{black} $R_E{}_i$ [Identify the primary emotions conveyed through $T_i$ and $EOR_i$ of the meme $\mathcal{X}_i$. ] 
% Additionally, analyze the emotions associated with the $EOR_i$. ]
\end{minipage}
}\\
\noindent
$R_E{}_i$ is the first prompt context, which infers the emotions-related rationale to provide the hidden cues for $\mathcal{X}_i$. $R_E{}_i=\text{argmax}(E|T_i, EOR_i)$ where $R_E{}_i$ is LLM-generated output text which explicitly mentions primary emotions E.  \\
\noindent
\textbf{Step 2. } After that, we ask LLM whether the meme is targeted towards women or not with the following template:\\
\noindent
\colorbox{gray!10}{
\begin{minipage}{0.95\columnwidth}
\color{black} $R_T{}_i$ [Based on the $T_i$ and the $EOR_i$ of the meme $\mathcal{X}_i$, provide a rationale for whether this meme targets women. Include specific elements that support this claim.]
\end{minipage}
}\\

\noindent
$R_T{}_i$ is the second prompt context, designed to extract a target-enriched rationale revealing cues of misogynous memes. It is defined as $R_T{}_i=\text{argmax}(T|T_i, EOR_i)$, where $R_T{}_i$ is the LLM-generated text that explicitly provides the rationale for whether the meme targets women. \\
\noindent
\textbf{Step 3. } Finally, to understand the broader context of a meme, we ask LLM to define the contextual information of the meme with the following template:\\
\noindent
\colorbox{gray!10}{
\begin{minipage}{0.95\columnwidth}
\color{black} $R_C{}_i$ [Given the text $T_i$ and the Entity-Object Relationships $EOR_i$ of the meme, provide the broader context $\mathcal{C}$ of meme $\mathcal{X}_i$.]
\end{minipage}
}\\
\noindent
Finally, $R_C{}_i$ is the third prompt context, aimed at uncovering contextual knowledge highlighting social cues associated with memes. It is defined as $R_C{}_i=\text{argmax}(C|T_i, EOR_i)$, where $R_C{}_i$ is the LLM-generated text explicitly outlining the meme's context C.
% \vspace{-0.1cm}
\subsection{Encoding of LLMs Generated Rationale }
To leverage the sequential and contextual information within the LLM-generated rationale $R_{r_i} = \{w_{1_i}, w_{2_i}, \ldots, w_{k_i}\}$, where $r \in \{e, t, c\}$ corresponds to emotion-rich, target-aware, and contextually-enriched rationale of meme sample $\mathcal{X}_i$, respectively, with varying word lengths $k\in\{l, m, n\}$, we employ the textual encoder of the pre-trained CLIP model:
\begin{equation}
% \vspace{-0.3cm}
(R_{E_i}, R_{T_i}, R_{C_i}) = \text{CLIP}(R_{e_i}, R_{t_i}, R_{c_i})~;
\end{equation}
\subsection{Enhancing CoT Reasoning via Cross-Attention}
We use three-layer hierarchical cross-attention to enable the interaction between the representations of rationales $(R_{E_i}, R_{T_i}, R_{C_i})$ before determining the final label $\hat{y}$.\\
\textbf{Emotive Multimodal Fusion (EMF):}
To derive an emotion-enriched multimodal representation of the meme $\mathcal{X}_i$, we calculate the cross-attention $H1_i$ between $M_i$ from Equation \ref{sumpool} and $R_{E_i}$. Initially, we perform linear transformations to obtain \emph{query} ($Q_{M_i} = M_i W_{M_q}$), \emph{key} ($K_{R_{E_i}} = R_{E_i} W_{E_k}$), and \emph{value} ($V_{M_i} = M_i W_{M_v}$) vectors for both the multimodal representation and the emotion-rich rationale using learned weight matrices ($W_{M_q}, W_{E_k}, W_{M_v}$):
\begin{equation}
% \small
H1_i = \text{softmax}\left(\frac{Q_{M_i} K_{R_{E_i}}^T}{\sqrt{d_k}}\right) V_{M_i}~;
\end{equation}
where $d_k$ is the dimension of the key vector. The final representation $HF1_i$ is obtained by adding $H1_i$ to the original multimodal representation $M_i$ through a residual connection and then applying layer normalization \cite{layer_2016}:
\begin{equation}\label{eq:EMFF}
HF1_i = \text{LayerNorm}(H1_i + M_i)~;
\end{equation}
\textbf{Target Insight Multimodal Representation (TIMR):}
To integrate the target-aware information of meme sample $\mathcal{X}i$ into the emotion-enriched representation obtained in Equation \ref{eq:EMFF}, we compute the cross-attention $H2_i$ between the emotive representation $HF1_i$ and the target-aware rationale $R_{T_i}$. We perform linear transformations to obtain \emph{query} ($Q_{HF1_i} = HF1_i W_{HF1_q}$), \emph{key} ($K_{R_{T_i}} = R_{T_i} W_{T_k}$), and \emph{value} ($V_{HF1_i} = HF1_i W_{HF1_v}$) vectors using learned weight matrices ($W_{HF1_q}, W_{T_k}, W_{HF1_v}$):
\begin{equation} %TIMR_i
% \small
H2_i = \text{softmax}\left(\frac{Q_{HF1_i} K_{R_{T_i}}^T}{\sqrt{d_k}}\right) V_{HF1_i}~;
\end{equation}
where $d_k$ is the dimension of the key vector. The final target-aware multimodal representation $HF2_i$ is obtained by adding $H2_i$ to the emotive representation $HF1_i$ through a residual connection and applying layer normalization:
\begin{equation}\label{eq:TIMRF}  %TIMRF_i
HF2_i = \text{LayerNorm}(H2_i + HF1_i)~;
\end{equation}
\textbf{Comprehensive Contextual Multimodal Insight (CCMI):}
To obtain a comprehensive contextual multimodal representation of meme sample $\mathcal{X}i$, we compute the cross-attention $H3_i$ between the target-aware representation $HF2_i$ and the context-aware rationale $R_{C_i}$. We perform linear transformations to obtain \emph{query} ($Q_{HF2_i} = HF2_i W_{HF2_q}$), \emph{key} ($K_{R_{C_i}} = R_{C_i} W_{C_k}$), and \emph{value} ($V_{HF2_i} = HF2_i W_{HF2_v}$) vectors using learned weight matrices ($W_{H2_q}, W_{C_k}, W_{H2_v}$):
\begin{equation}
% \small
H3_i = \text{softmax}\left(\frac{Q_{HF2_i} K_{R_{C_i}}^T}{\sqrt{d_k}}\right) V_{H2_i}~;
\end{equation}
where $d_k$ is the dimension of the key vector. The final comprehensive representation $HF3_i$ is obtained by adding $H3_i$ to the target-aware representation $HF2_i$ through a residual connection and applying layer normalization:
\begin{equation}\label{eq:CCMI}
HF3_i = \text{LayerNorm}(H3_i + HF2_i)~;
\end{equation}
\subsection{Network Training}
We use a singular feed-forward neural net (FFN) with softmax activation, which takes the Comprehensive Contextual Multimodal representation $(HF3_i)$ in Equation \ref{eq:CCMI} as input and outputs class for misogynous meme identification, shown in the following Equation \ref{loss_new}:
\begin{equation}\label{loss_new}
% \small
\begin{aligned}
       \hat{y}_{t}=P(Y_i|HF3_{i},W,b)=\mathnormal{\text{softmax}}(HF3_{i}W_i+b_i);
       \end{aligned}
       % \vspace{-0.3cm}
\end{equation}
The proposed classifier is trained using cross-entropy loss: 
\begin{equation}\label{ov_loss}
% \small
 \mathcal{L}_{1}= -\sum[y_{t}\log \hat{y}_{t}+(1-y_{t})\log (1-\hat{y}_{t})]~;
\end{equation}
\textbf{Reasoning Revising with Supervised Contrastive Learning Loss:}
In addition to cross-entropy loss, we incorporate supervised contrastive loss (SCL) to enhance the CoT-based learning and provide empirical evidence of its effectiveness in learning cultural diversity-enriched representations for a more robust classifier \cite{li-etal-2023-prompt,shen2021contrastive}. This loss component encourages well-separated representations for the misogynous meme identification task, creating equitable representations and correct predictions. All three multimodal representations that enhance the CoT reasoning,i.e., ($HF1_i$,$HF2_i$,$HF3_i$ in Equation \ref{eq:EMFF}, \ref{eq:TIMRF}, and \ref{eq:CCMI}) and multimodal representation $M_i$, are assumed to capture similar contexts for a given meme $\mathcal{X}_i$. During training, these representations are aligned within the same semantic space, enabling effective utilization through contrastive learning.
\begin{equation}\label{eq:con}
\scriptsize
% \small
\begin{aligned}
     \mathcal{L}_{Em}=-\log \frac{\exp \left(\operatorname{sim}\left(\boldsymbol{HF1_i}, \boldsymbol{M_i}\right) / \tau\right)}{\sum_{l=1[l \neq i]}^{2 N}\exp \left(\operatorname{sim}\left(\boldsymbol{HF1_i}, \boldsymbol{M_l}\right) / \tau\right)}~;\\
    \mathcal{L}_{Ta}=-\log \frac{\exp \left(\operatorname{sim}\left(\boldsymbol{HF2_i}, \boldsymbol{M_i}\right) / \tau\right)}{\sum_{l=1[l \neq i]}^{2 N}\exp \left(\operatorname{sim}\left(\boldsymbol{HF2_i}, \boldsymbol{M_l}\right) / \tau\right)}~;\\
     \mathcal{L}_{Co}=-\log \frac{\exp \left(\operatorname{sim}\left(\boldsymbol{HF3_i}, \boldsymbol{M_i}\right) / \tau\right)}{\sum_{l=1[l \neq i]}^{2 N}\exp \left(\operatorname{sim}\left(\boldsymbol{HF3_i}, \boldsymbol{M_l}\right) / \tau\right)}~;\\
       \end{aligned}
\end{equation}
where, $sim$ is the cosine-similarity, $N$ is the batch size, and $\tau$ is the temperature to scale the logits.
Therefore, the overall loss $\mathcal{L}_{F}$ is a weighted sum of the cross-entropy loss $\mathcal{L}_{1}$ in Equation \ref{ov_loss}, and these contrastive losses ($\mathcal{L}_{Em}, \mathcal{L}_{Ta}, \mathcal{L}_{Co}$) in Equation \ref{eq:con}. The weights  ($\alpha$, $\beta$,$\gamma$, and $\theta$) control the relative importance of each loss. \\
\begin{equation}\label{eq:overall}
% \small
    \mathcal{L}_{F} = \alpha \cdot \mathcal{L}_{1} + \beta \cdot \mathcal{L}_{Em} +
    \gamma \cdot \mathcal{L}_{Ta} + \theta \cdot \mathcal{L}_{Co}~;
        % \end{aligned}
    % \vspace{-0.5cm}
\end{equation}
\begin{table*}[!htp]\centering
\adjustbox{width=0.7\linewidth}{\begin{tabular}{lrrrrrrrrrrrrrr}\toprule
\multicolumn{2}{c}{\multirow{3}{*}{\textbf{Models}}} &\multirow{3}{*}{\textbf{Text}} &\multirow{3}{*}{\textbf{Image}} &\multicolumn{6}{c}{\textbf{MAMI}} &\multicolumn{4}{c}{\textbf{MIMIC}} \\\cmidrule{5-14}
& & & &\multicolumn{3}{c}{\textbf{Dev}} &\multicolumn{3}{c}{\textbf{Test}} &\multicolumn{4}{c}{\textbf{Test}} \\\cmidrule{5-14}
& & & &\textbf{P} &\textbf{R} &\textbf{M-F1} &\textbf{P} &\textbf{R} &\textbf{M-F1} &\textbf{P} &\textbf{R} &\textbf{M-F1} &\textbf{W-F1} \\\cmidrule{1-14}
\multirow{17}{*}{\textbf{Baseline}} &$L_{FT} (1)$ &\checkmark & &56.23 &56.69 &56.47 &44.7 &47.9 &46.2 &47.39 &46.93 &47.16 &47.29 \\
&BERT(2) & &\checkmark &63.29 &71.81 &67.28 &58.0 &50.9 &54.2 &61.45 &61.06 &61.25 &61.26 \\
&LaBSE (3) &\checkmark & &63.59 &61.99 &63.72 &49.4 &54.2 &51.6 &63.59 &61.39&62.48 &62.66 \\
&VGG\-19 (4) & & \checkmark & 64.29	&60.79&62.49 &47.40 &49.40 &48.38 &44.48 &42.35 &43.39 &43.84 \\
&ViT (5) & &\checkmark & 69.21&	67.36 &68.27 &54.30 &52.40 &53.37 &49.99 &48.91 &49.45 &49.29 \\ \cmidrule{2-14}
&\textbf{Early Fusion} &\textbf{} &\textbf{} & & & & & & & & & & \\
&(1)+(4) & &\checkmark &72.60 &62.52 &67.19 &52.5 &47.0 &49.6 &52.39 &50.38 &51.37 &51.2 \\
&(2)+(4) &\checkmark &\checkmark &58.19 &64.48 &61.18 &54.4 &51.3 &52.7 &64.29 &62.49 &63.38 &63.24 \\
&(2)+(5) &\checkmark &\checkmark &70.81 &64.09 &68.27 &53.48 &59.29 &56.21 &69.49 &67.97 &68.72 &68.49 \\
&(3)+(5) &\checkmark &\checkmark &69.09 &61.93 &65.28 &55.93 &51.19&53.0 &63.85 &63.94 &63.89 &63.91 \\\cmidrule{2-14}
&\textbf{Pretrained Model} &\textbf{} &\textbf{} & & & & & & & & & & \\
&LXMERT &\checkmark &\checkmark &78.94 &69.45 &73.88 &69.01 &65.18 &65.9 &66.03 &61.39 &63.63 &63.21 \\
&MMBT &\checkmark &\checkmark &73.60 &69.09 &71.27 &56.4 &49.0 &52.4 &68.39 &65.91 &67.13 &67.39 \\
&VisualBERT &\checkmark &\checkmark &81.03 &77.79 &79.38 &78.2 &61.2 &68.7 &73.98 &70.39 &72.15 &72.38 \\
&BLIP &\checkmark &\checkmark &70.95 &68.28 &69.58 &62.39 &53.39 &57.54 &74.39 &72.39 &73.38 &73.74 \\
&ALBEF &\checkmark &\checkmark &72.30 &70.98 &71.62 &59.2 &53.5 &56.1 &71.21 &69.38 &70.28 &70.13 \\
&*\textbf{CLIP\_MM} &\checkmark &\checkmark &\textbf{85.3} &\textbf{83.4} &\textbf{84.3} &\textbf{75.4} &\textbf{69.2} &\textbf{72.1} &\textbf{76.39} &\textbf{74.05} &\textbf{75.24} &\textbf{75.25} \\ \midrule
\multirow{4}{*}{\rotatebox{90}{\textbf{Prompt-based}}}&\multicolumn{6}{c}{\textbf{CLIP\_MM}} \\
&+ ChatGPT &\checkmark &\checkmark &85.89 &83.99 &84.98 &80.0 &69.3 &74.2 &76.71 &74.59 &75.63 &75.34 \\
&+ GPT 4 &\checkmark &\checkmark &87.11 &84.81 &85.93 &75.5 &71.3 &72.3 &76.47 &72.43 &74.39 &74.12 \\
&+ Llama &\checkmark &\checkmark &83.70 &81.29 &82.46 &77.83 &69.40 &73.38 &78.01 &73.97 &75.94 &75.75 \\
&\textbf{+ Mistral (Ours)} &\checkmark &\checkmark &\textbf{88.80} &\textbf{84.76} &\textbf{86.72} &\textbf{81.20} &\textbf{72.70} &\textbf{76.94} &\textbf{78.15} &\textbf{75.39} &\textbf{76.75} &\textbf{76.35 }\\ \midrule
\multirow{6}{*}{\rotatebox{90}{\textbf{CoT-based}}}&\multicolumn{6}{c}{\textbf{CLIP\_MM}}\\
% && & & & && & & & & & & \\
&\textbf{+}ChatGPT &\checkmark &\checkmark &86.20 &84.40 &85.29 &81.0 &76.0 &77.0 &78.69 &76.34 &77.49 &77.41 \\
&\textbf{+} GPT4 &\checkmark &\checkmark &89.52 &85.20 &87.38 &71.9 &70.8 &71.4 &75.16 &73.39 &74.26 &74.21 \\
&$+ Llama$ &\checkmark &\checkmark &91.38 &86.28 &88.85 &77.50 &76.40 &76.98 &77.17 &75.10 &76.12 &76.91 \\ \cmidrule{2-14}
&$\textbf{M3Hop-CoT}_{Mistral}$ &\multirow{2}{*}{\checkmark} &\multirow{2}{*}{\checkmark} &\multirow{2}{*}{\textbf{96.39}}&\multirow{2}{*}{\textbf{87.59}}&\multirow{2}{*}{\textbf{91.75}} &\multirow{2}{*}{\textbf{82.38} }&\multirow{2}{*}{\textbf{78.29}} &\multirow{2}{*}{\textbf{80.28}} &\multirow{2}{*}{\textbf{80.29}}&\multirow{2}{*}{\textbf{78.98}}&\multirow{2}{*}{\textbf{79.63}} &\multirow{2}{*}{\textbf{79.41}} \\
&\textbf{(Proposed)} & && & &&& & & &&& \\
\bottomrule
\end{tabular}}
\caption{Results from the proposed model and the various baselines on the MAMI and MIMIC datasets. Here, the bolded values indicate maximum scores. Here, T: Text, I: Image, M-F1: Macro F1, and W-F1: weighted F1-score. * represents the best-performing baseline model. We observe that the performance gains are statistically significant with p-values (<0.0431) using a t-test, which signifies a 95\% confidence interval.}\label{tab:result}
% \vspace{-0.7cm}
\end{table*}
\section{Results Analysis}
In this section, we present the results of our comparative analysis, which examines the baseline models \footnote{Details of the baseline models are given in the Appendix Section \ref{sec:app_baselines}}, LLM-based models, our proposed model, and their respective variations for misogynous meme identification tasks\footnote{Additional details of experimental setups and hyperparameters explored are provided in the Appendix Section \ref{app:im_det}}. 
We use the macro-F1 (F1) score on both the dev and test sets as the preferred metrics to measure this. 
\subsection{Model Results and Comparisons}
\textbf{Models Notation:} CLIP\_MM: This is the CLIP-based classifier. M3Hop-CoT: Proposed scene graph with CoT-based model with emotion, target, and context-aware prompts. M3Hop-CoT$^{-E}$: Proposed model without emotion-aware prompt,  M3Hop-CoT$^{-T}$: Proposed model without target-aware prompt, M3Hop-CoT$^{-C}$:  Proposed model without context-aware prompt, M3Hop-CoT$^{-SG}$: This model is trained solely with all the CoT based modules, excluding the scene graph, M3Hop-CoT$^{E}$: Proposed model with only emotion-aware prompt, M3Hop-CoT$^{T}$: Proposed model with only target-aware prompt, M3Hop-CoT$^{C}$: Proposed model with only context-aware prompt and M3Hop-CoT$^{-\mathcal{L}_k}$ where k$\in\{Em, Ta, Co\}$: Proposed model without respective k$_{th}$ loss.\\ 
% \noindent
\subsubsection{Results on MAMI Dataset}
\textbf{Comparison with Baseline Models:} Table \ref{tab:result} presents the performance of various baseline models on the task of misogynous meme identification. Notably, our CLIP-based baseline classifier (CLIP\_MM) achieves superior performance with an F1 score of 73.84\% on both the dev and test sets, serving as the foundation for our proposed method. We also observed that multimodal baselines give better results than unimodal ones. Furthermore, our proposed model, M3Hop-CoT, surpasses all other baseline models in terms of F1 scores for both the dev and test datasets. It shows the robustness of our proposed model for such a challenging task. \\
\noindent
\textbf{Comparison with LLMs:} 
When extending the CLIP\_MM with a prompt-based approach, Mistral LLM surpasses other LLMs by achieving a $\backsim2$\% increment on the dev set, whereas $\backsim4$\% higher F1-score on the test set, establishing a strong foundation for subsequent CoT-based methods. Moreover, when implementing the CoT-based approach across various LLMs, M3Hop-CoT, which incorporates Mistral LLM, consistently outperforms other CoT-based models. It validated the robustness of the proposed model, which understands the hidden complex cues of any meme by means of their hidden emotions, target, and contextual information (A detailed discussion about the comparison of only prompt-based models with CoT-based models with different LLMs is given in Appendix Section \ref{sec:app_llm}).\\

\begin{table}[!h]\centering
\adjustbox{width=0.8\linewidth}{\begin{tabular}{lrrrrrrr}\toprule
\multirow{3}{*}{\textbf{Models}} &\multirow{3}{*}{\textbf{Text}} &\multirow{3}{*}{\textbf{Image}} &\multicolumn{4}{c}{\textbf{Macro F1-score}} \\\cmidrule{4-7}
& & &\multicolumn{2}{c}{\textbf{MAMI}} &\multicolumn{2}{c}{\textbf{MIMIC}} \\\cmidrule{4-7}
& & &\textbf{dev } &\textbf{test} &\textbf{M-F1 } &\textbf{W-F1} \\\cmidrule{1-7}
\textbf{M3Hop-CoT (Ours)} &\checkmark &\checkmark &\textbf{91.75} &\textbf{80.28} &\textbf{79.63}&  \textbf{79.41 }\\ \midrule
 $M3Hop-CoT^{-E}$ &\checkmark &\checkmark &86.83 &76.3 & 73.74& 73.01\\
$M3Hop-CoT^{-T}$ &\checkmark &\checkmark &86.92 &75.1 &75.37 &74.92 \\
$M3Hop-CoT^{-C}$ &\checkmark &\checkmark &85.92 &75.3 & 73.91& 73.24\\
$M3Hop-CoT^{-SG}$ &\checkmark &\checkmark &84.21 &73.9 & 72.35& 72.47\\
$M3Hop-CoT^{E}$ &\checkmark &\checkmark &82.99 &70.2 & 69.28& 70.14\\
$M3Hop-CoT^{T}$ &\checkmark &\checkmark &84.21 &73.2 & 73.58& 73.76\\
$M3Hop-CoT^{C}$ &\checkmark &\checkmark &84.38 &71.2 &75.86 &75.97 \\
$M3Hop-CoT^{-\mathcal{L}\_{Em}}$ &\checkmark &\checkmark &89.29 &76.2 & 77.94& 77.05\\
$M3Hop-CoT^{-\mathcal{L}\_{Ta}}$ &\checkmark &\checkmark &88.73 &77.0 & 75.62&75.33 \\
$M3Hop-CoT^{-\mathcal{L}\_{Co}}$ &\checkmark &\checkmark &88.28 &77.9 &77.95 &77.01 \\
\bottomrule
\end{tabular}}\caption{Ablation Study: Role of different modules in our proposed model. We observe that the performance gains are statistically significant with p-values ($<$0.0553) using a t-test, which signifies a $95$\% confidence interval.}\label{tab:ablation}
% \vspace{-0.5cm}
\end{table}
\noindent
\subsubsection{Results on MIMIC Dataset}
To show the robustness of our proposed model in another language, in Table \ref{tab:result}, we have shown the results on the MIMIC dataset, which is in Hindi-English code-mixed. Our proposed model follows a behavior similar to the MAMI dataset (outperforming by more than $\backsim4-5$\% from CLIP\_MM), whereas CoT-based LLM is not only leveraging the language-related dependency but also performing superbly by utilizing the different cultural-based hidden cues of the dataset (A detailed analysis of results on this dataset is provided in the Appendix Section \ref{sec:MIMIC}).
\subsubsection{Ablation Study} To assess our proposed architecture, we created several multimodal variants of our proposed model M3Hop-CoT by training it on MAMI and MIMIC datasets, as shown in Table \ref{tab:ablation}, which allows us to evaluate the contribution of each component to the model's overall performance. M3Hop-CoT emerged as the most effective model, achieving a significant increase of 6-7\% in F1 scores for both development and test sets. Additionally, incorporating SCL further enhanced M3Hop-CoT's performance, as evidenced by the impact of each loss component. The model's superior performance is attributed to its balanced use of textual and visual modalities, integration of entity-object relationships, and leveraging key factors such as emotion, target, and context-enriched LLM-generated rationales. M3Hop-CoT effectively captures the semantic relationships between objects in the meme, which is crucial for identifying misogynous content.
\subsection{Detailed Analysis}
\subsubsection{Result Analysis with Case Study}
Using Appendix Figure \ref{fig:detailed}, we qualitatively analyze our proposed framework through the predictions obtained from the baseline CLIP\_MM and our proposed model M3Hop-CoT. For meme sample (a) with the text ``I WAS BROUGHT UP TO NEVER HIT A WOMAN. SHE'S NO WOMAN," and an image showing a slap and a woman, ``CLIP\_MM," classified it as non-misogynous. In contrast, our model M3Hop-CoT correctly classified it as misogynous using a CoT-based rationale from an LLM with multi-hop reasoning. While CLIP\_MM slightly preferred text (as depicted by T= 13.85) over visuals (V= 11.27), M3Hop-CoT provided a balanced contribution by considering both text and visuals and context. It is evident in GradCAM, where M3Hop-CoT distinctly highlights both the slap and the woman, unlike CLIP\_MM, which fails to concentrate on these critical elements.
Similarly, the meme sample (b) conveys the disrespect towards women using domestic violence. The LLM's generated rationale offers insight into the meme's intended message. Once again, CLIP\_MM struggles to accurately classify the meme, whereas ``M3Hop-CoT" correctly identifies it as misogynous. M3Hop-CoT effectively recognizes the sarcastic nature of memes by underlying emotions, target, and context, showcasing their ability to understand the meme's subtleties.
In example (c), the meme, which compares a woman to a pig, is identified as misogynous. The CLIP\_MM fails to classify it correctly, focusing only on the words "EX WIFE/FOLLOW/ WEDDING PHOTOS" and missing the image's subtle cues. In contrast, ``M3Hop-CoT" accurately detects its misogynous nature by considering both modalities and integrating contextual knowledge through multimodal reasoning. Enhanced by CoT prompting and EoRs, M3Hop-CoT provides a more comprehensive analysis and outperforms baseline models in recognizing misogynous content (Similar qualitative analysis for the MIMIC dataset is shown in the Appendix Section \ref{sec:mimic_quali}.)
\subsubsection{Result Analysis for Cultural Diversity}
In the Appendix Figure \ref{fig:cultural}, we present three illustrative examples from the MAMI dataset, showcasing how M3Hop-CoT leverages cultural knowledge from diverse demographics. The model better recognizes misogyny by incorporating emotional cues, target identification, and context in a CoT framework. Each example in the figure delves into different cultural references. These include historical beliefs surrounding the Church and women's roles in the 1500s (c.f. example (i)), comparisons between women and witches within Japanese mythology ( c.f. example (ii)), and Christian interpretations of the Bible's teachings (c.f. example (iii)). Notably, CLIP\_MM fails to grasp the underlying misogynistic connotations within these examples. Conversely, our proposed model effectively utilizes these cultural references, leading to accurate predictions of misogynistic labels.
\subsubsection{Quantitative Analysis with Error Rates}
We illustrate the impact of various M3Hop-CoT model variants on test error rates in the Appendix Figure \ref{fig:miss_error}. CLIP\_MM model exhibits the highest error rate, highlighting the necessity of LLMs for such complex tasks. Models like M3Hop-CoT$^{-E}$, M3Hop-CoT$^{-T}$, and M3Hop-CoT$^{-C}$, lack emotion, target, and context-aware prompts, respectively, have higher error rates than the proposed M3Hop-CoT, indicating the importance of these components. Additionally, M3Hop-CoT$^{-SG}$, excluding the scene graph module, shows an increased error rate, emphasizing the significance of visual semantics. Models M3Hop-CoT$^{E}$, M3Hop-CoT$^{T}$, and M3Hop-CoT$^{C}$, focusing on individual rationale, demonstrate that a balanced approach is essential for optimal performance. The M3Hop-CoT model achieves the lowest error rates, demonstrating its superior ability to identify such memes. 
% \subsection{Transferability of the Proposed Model} 
\subsection{Generalibity of the Proposed Model} 
To demonstrate the adaptability of our proposed architecture, M3Hop-CoT, we assess its performance across three English benchmark datasets: Hateful Memes, Memotion2, and the Harmful dataset (c.f. Table \ref{tab:result_gen}). This evaluation validates the generalizability of our architecture, demonstrating its effectiveness not only in misogynous tasks but also in various benchmark datasets and tasks (See the detailed discussion in Appendix Section \ref{sec:detail_generic_data}).
\begin{table}[!htp]\centering
\scriptsize
\adjustbox{width=\linewidth}{\begin{tabular}{lrrrrrrr}\toprule
\multirow{3}{*}{\textbf{Models}} &\multicolumn{2}{c}{\multirow{2}{*}{\textbf{Modality}}} &\textbf{Memotion2} &\multicolumn{2}{c}{\textbf{Hateful}} &\textbf{Harmful} \\\cmidrule{4-7}
&\textbf{T} &\textbf{I} &\textbf{F1$\uparrow$} &\textbf{F1$\uparrow$} &\textbf{AUC$\uparrow$} &\textbf{F1$\uparrow$} \\\cmidrule{1-7}
$FasterRCNN$ & &\checkmark &48.9 &38.81 &59.97 &65.9 \\
BERT &\checkmark & &50.01 &58.41 &67.92 &77.92 \\
ViT & &\checkmark &51.17 &--- &--- &67.88 \\
Late-Fusion &\checkmark &\checkmark &51.4 &64.40 &72.51 &78.50 \\
$MMBT$ &\checkmark &\checkmark &52.1 &58.29 &76.77 &80.2 \\
$Visual BERT COCO$ &\checkmark &\checkmark &50.86 &59.28 &73.85 &86.1 \\
$ALBEF$ &\checkmark &\checkmark &50.8 &--- &--- &87.5 \\
$ViLBERT$ &\checkmark &\checkmark &49.92 &52.60 &76.32 &85.83 \\
VisualBERT &\checkmark &\checkmark &51.06 &67.46 &74.63 &84.57 \\
$UNITER$ &\checkmark &\checkmark &52.7 &61.66 &60.02 &61.66 \\
$LXMERT$ &\checkmark &\checkmark &52.3 &69.45 &76.15 &69.45 \\
$^\phi$SOTA &\checkmark &\checkmark &$^\varphi$55.17 &$^\gamma$66.71 &73.43 &$^\gamma$89.0 \\
DisMultiHate &\checkmark &\checkmark &50.57 &63.31 &75.97 &84.57 \\
Momenta &\checkmark &\checkmark &50.9 &66.71 &73.43 &88.3 \\
PromptHate &\checkmark &\checkmark &50.89 &71.22 &77.07 &89.0 \\ \midrule
CLIP\_MM (Full-Train) &\checkmark &\checkmark &48.4 &53.22 &75.98 &82.9 \\
CLIP\_MM+GPT4 (Full-Train) &\checkmark &\checkmark &56.39& 62.18 &77.13 &85.64 \\
CLIP\_MM+ ChatGPT (Full-Train) &\checkmark &\checkmark &55.74& 60.39& 76.21&86.29 \\
CLIP\_MM+Llama (Full-Train)&\checkmark &\checkmark &56.23&59.63 & 78.29& 88.29\\
CLIP\_MM+Mistral (Full-Train) &\checkmark &\checkmark &57.75& 65.58 &79.02 &88.75 \\
\midrule
M3Hop-CoT (Zero-Shot)&\checkmark &\checkmark &\textbf{53.47} &\textbf{78.36 }&\textbf{79.93} &\textbf{85.38} \\
M3Hop-CoT (Full-Train)&\checkmark &\checkmark &\textbf{59.95} &\textbf{79.24 }&\textbf{83.29} &\textbf{91.01} \\
\bottomrule
\end{tabular}}
\caption{Results from the proposed model and the various baselines on the Memotion2, Hateful Memes, and Harmful Memes datasets. $\Phi$: SOTA model on respective datasets. $\varphi$ by \cite{memotion2} for Memotion2,  $\gamma$ by \cite{cao-etal-2022-prompting} for Hateful meme, and Harmful Memes. We observe that the performance gains are statistically significant with p-values ($<$0.05) using a t-test, which signifies a $95$\% confidence interval.}\label{tab:result_gen}
\end{table}
\subsection{Comparison with State-of-the-Art Models}
Table \ref{tab:comparison} presents a detailed comparison between M3Hop-CoT and other state-of-the-art (SOTA) models. In the MAMI task, M3Hop-CoT surpasses existing SOTA. Despite PromptHate achieving high accuracy on the MAMI dataset, it struggles with contextual knowledge, leading to modality-specific biases. Another model, Multimodal-CoT, attempts to leverage multimodal features with LLM but lacks essential psycholinguistic factors that our model incorporates, such as emotions, target awareness, and contextual information (c.f. Figure \ref{fig:llm_prediction} ). Our M3Hop-CoT model outperforms, mainly due to its use of EoRs and the above psycholinguistic factors. 
\begin{table}[!h]\centering
% \vspace{-0.5cm}
\adjustbox{width=0.7\linewidth}{\begin{tabular}{lrrr}
\toprule
\multirow{2}{*}{\textbf{Models}} & \multicolumn{2}{c}{\textbf{Macro-F1$\uparrow$} }\\
&\textbf{dev} &\textbf{test}\\
\midrule
$^\Psi$\citet{zhang-wang-2022-srcb} &83.4& 77.6 \\ %77.6, 62.49
DisMultiHate \cite{10.1145/3474085.3475625}& 67.24 & 61.89\\
Momenta \cite{pramanick-etal-2021-momenta-multimodal}&72.81&68.29 \\
MMBT \cite{kiela2019supervised} &74.8&68.93 \\
PromptHate \cite{cao-etal-2022-prompting} & 79.98 & 73.28\\
Multimodal-CoT \cite{zhang2023multimodal}& 82.98 & 72.19\\
\citet{kumari-etal-2024-eacl}& 79.59 & ---\\   \midrule
\textbf{M3Hop-CoT}  & \textbf{91.75} & \textbf{80.28} \\
\bottomrule
\end{tabular}}
\caption{Comparison of our proposed model with the existing SOTA models, $\Psi$ is the SOTA on MAMI Dataset}\label{tab:comparison}
\end{table}
% Another contributing factor the model benefits from SCL,  
\begin{figure}[t!]
\centering
\includegraphics[width=\linewidth]{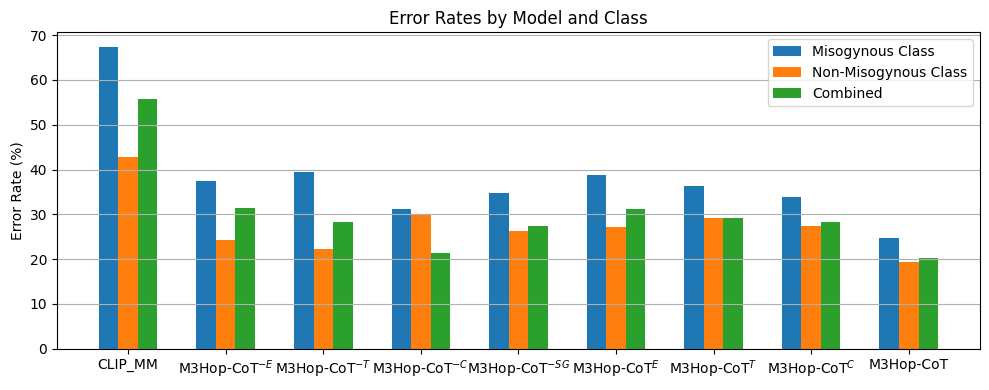}
\caption{Misclassification rate comparison between proposed model \textbf{M3Hop-CoT} and their various variants\label{fig:miss_error}}
\end{figure}
\begin{figure}[t!]
\centering
\includegraphics[width=\linewidth]{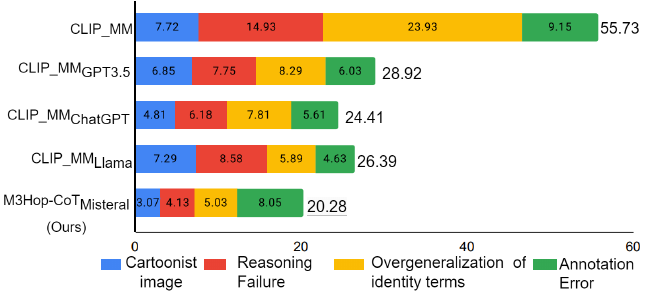}
\caption{ Categorization of error analysis (\%) of proposed model \textbf{M3Hop-CoT} and other SOTA models\label{fig:miss_rate}}
\end{figure}
\subsection{Error Analysis }\label{sec:error}
Despite its high performance, our proposed model occasionally misclassifies memes in following scenarios: 
\textbf{(i) Cartoonist image:} In certain scenarios, M3Hop-CoT overlooks the extracted rationale from CoT LLMs and solely concentrates on the image featuring cartoon characters, leading to a misclassification of the meme as ``Non-misogynous (c.f. Appendix Figure \ref{fig:error_rationale} (a))." \textbf{(ii) Reasoning Failure:}  M3Hop-CoT sometimes struggle to produce accurate rationales using LLMs due to the implicit nature of memes (c.f. Appendix Figure \ref{fig:error_rationale} (b)), such as failing to recognize external references (e.g., the significance of grey sweatpants). \textbf{(iii) Overgeneralization of identity terms:} M3Hop-CoT overgeneralize specific ``identity terms (e.g., the presence of the word `SANDWICH')," leading to a misclassify a meme as ``Misogynous" based solely on these words while disregarding other information such as images and rationales extracted by LLMs (c.f. Appendix Figure \ref{fig:error_rationale} (c)). \textbf{(iv) Annotation Error:} In our analysis, we encountered situations where our proposed model accurately predicted the correct label for a given sample. However, due to the problematic annotation issues, misclassification happens (c.f. Appendix Figure \ref{fig:error_rationale} (d)). (More detailed error analyses are discussed in Appendix Section \ref{sec:appen_error}). 
\section{Conclusion}
In conclusion, our work introduces a novel approach for detecting misogynous content in memes, leveraging the power of LLMs with CoT reasoning. Our proposed model, \textit{M3Hop-CoT}, integrates multimodal information and employs a three-step reasoning process to effectively capture memes' emotional, target-oriented, and contextual nuances. By incorporating scene graphs, we enhance the model's ability to understand the visual aspects of memes. Our results demonstrate that M3Hop-CoT outperforms existing SOTA models, significantly improving F1 scores on both dev and test sets.  
In the future, we could explore extending our approach to other forms of online content and integrating additional modalities to enhance the model's effectiveness.
\section*{Limitations}
In Section \ref{sec:error}, we discussed a few limitations of our proposed model. Despite its strengths, our model encounters difficulties in accurately detecting misogynous memes, especially when the images are cartoonish or when the misogynous references are subtle and require nuanced reasoning. These challenges highlight areas for further refinement and improvement. Understanding these limitations is crucial for advancing our model's capability to identify misogynous content more effectively in future iterations (See a detailed future discussion in the Appendix Section \ref{sec:fututre_work}).
\section*{Ethics Statement}
\textbf{Broader Impact:} The broader impact of this work is significant in the field of misogynous meme identification. This research promotes a safer and more respectful online environment by developing advanced techniques for detecting misogynous content. Our proposed model, M3Hop-CoT, can help reduce the prevalence of harmful content, fostering a more inclusive and peaceful digital community. Addressing the issue of detecting misogynous memes is essential for promoting equality and fostering peace and justice. We create a more inclusive and fair online environment by developing methods to identify such internet memes. This effort also supports the principle by ensuring that marginalized and vulnerable genders are included in development initiatives. 
However, it is important to acknowledge the ongoing discussion of automated content moderation and potential biases within such systems. We will explore techniques to ensure fairness, transparency, and accountability in future work in such models (See a detailed future discussion in the Appendix Section \ref{sec:fututre_work}).
\linebreak
\textbf{Intended Use:} This research is intended to advance the detection of misogynous content on social media, aiming to improve the experiences of social media users, content moderators, and the broader online community. By enhancing the ability to identify and moderate such content, we hope to contribute positively to safer online interactions.
\linebreak
\textbf{Misuse Potential:} The dataset utilized in this study includes memes with slur words and offensive images, which are included solely for understanding and analyzing the dataset. It is important to clarify that our use of such content is strictly for research, and we do not intend to harm any individual or group. We emphasize the ethical use of our findings and the importance of handling sensitive content with care.
\section*{Acknowledgements}
The research reported in this paper is an outcome of the project ``\textbf{HELIOS: Hate, Hyperpartisan, and Hyperpluralism Elicitation and Observer System,}" sponsored by Wipro AI Labs, India.
% \bibliography{custom}

\appendix
\section{Detailed Results Analysis on MIMIC Dataset}\label{sec:MIMIC}
In Table \ref{tab:result}, we have mentioned the results of our proposed model and several baseline models for the MIMIC dataset. Notably, the baseline model CM\_CLIP performed better than other baselines, showcasing the efficiency of the pre-trained CLIP model for multimodal data. It is performing better than other baselines, with more than $\backsim4$\% increment on the test dataset in terms of macro-F1 and weighted F1-scores. Now, moving towards using LLMs, it replicates results similar to those of the MAMI dataset. Although Llama's performance on the MIMIC dataset is better than the MAMI dataset, Mistral LLM is again providing better context than other LLMs, resulting in an increment of $\backsim3$\%. For the proposed model 
\subsection{Qualitative Analysis of the MIMIC Dataset}\label{sec:mimic_quali}
In Appendix Figure \ref{fig:detailed_MIMIC}, we present a qualitative analysis comparing the performance of our proposed model, M3Hop-CoT, with the baseline model, CM\_CLIP, on the MIMIC dataset. Sample (i) depicts a meme intended to degrade women through prejudice. CM\_CLIP fails to understand the underlying prejudices. In contrast, M3Hop-CoT, by leveraging its ability to understand emotions, targeted information, and context, correctly identifies the misogynistic nature of this meme. Similarly, samples (ii) and (iii) showcase memes designed to humiliate women by referencing a specific Indian context. M3Hop-CoT demonstrates human-like comprehension of the subtle humiliation conveyed within these memes, leading to accurate predictions of the misogynistic label. These findings highlight the effectiveness of M3Hop-CoT in identifying misogyny compared to the baseline model.
\section{Experiments}
\subsection{Baseline Models}\label{sec:app_baselines}
To compare the performance of our proposed model with some existing state-of-the-art models, we create several baseline models.  
\subsection{Unimodal Systems}
For the unimodal setting, we implement the following variants of the baseline models: 
\begin{enumerate}[nolistsep,noitemsep]
    \item \textbf{LSTM with FastText-based Embedding ($L_{FT}$)}: We utilize LSTM (Long Short-Term Memory) \cite{hochreiter1997long} networks combined with FastText embeddings \cite{joulin2016bag} to leverage both sequential processing capabilities and enriched word vector representations. 
    \item \textbf{BERT (Bidirectional Encoder Representations from Transformers):} Next, we leverage the BERT model \cite{pires-etal-2019-multilingual} to extract contextually rich feature representations from meme text.
    \item \textbf{LaBSE (Language-agnostic BERT Sentence Embedding): } We utilized the LaBSE \cite{DBLP:journals/corr/abs-2007-01852} model to obtain high-quality language-agnostic text embeddings of the meme text. 
    \item \textbf{VGG-19:} This VGG-19 \cite{Simonyan15} architecture is included to capture visual features from meme images. VGG-19 is highly effective in extracting intricate patterns and textures from visual data, which are crucial for analyzing image-based content.
    \item \textbf{Visual Transformer (ViT):} The ViT \cite{DBLP:journals/corr/abs-2010-11929} model applies the principles of transformers for image recognition. This model segments images into patches and processes them sequentially, enabling the capture of global dependencies across the entire image.
\end{enumerate}
After feature extraction from each model, the resulting feature vectors are processed through a softmax function for final prediction. 
\subsection{Multimodal Systems}
\textbf{Early Fusion:}
In our early fusion approach, we leverage the strength of combining textual and visual features at an initial processing stage by concatenating them to enhance the model's understanding of misogynous context. 
\begin{enumerate}[nolistsep,noitemsep]
    \item \textbf{$\textbf{L}_{\textbf{FT}}$+VGG:} Combines LSTM with FastText-based embedding for text and VGG19 for image features, integrating rich textual embeddings with image features.
    \item \textbf{BERT+VGG: } This model utilizes BERT for its superior text features and VGG-19 for robust image feature extraction.
    \item \textbf{BERT+ViT:} This model concatenates BERT's contextual understanding of the text with a visual transformer for image features for the final prediction.
    \item \textbf{LaBSE+ViT:} In this model, we paired Language-agnostic BERT Sentence Embeddings with a Visual Transformer for processing multilingual text alongside complex image data.
\end{enumerate}
\textbf{Pre-trained Models:}
To get a better multimodal representation, we employ  different pre-trained models that are specifically designed for handling complex multimodal data:
\begin{enumerate}[nolistsep,noitemsep]
    \item \textbf{LXMERT} \cite{tan-bansal-2019-lxmert}: This model is specifically tailored for learning cross-modality representations and has shown exceptional performance on tasks that require joint understanding of text and image content.
    \item \textbf{VisualBERT} \cite{zhou-etal-2022-dd-tig}: A variant of BERT incorporating visual features into the BERT architecture, enhancing its applicability to scenarios where visual context is crucial.
    \item \textbf{MMBT} (Supervised Multimodal Bitransformers) \cite{kiela2019supervised}: MMBT integrates information from heterogeneous sources (text and image) using transformer architectures, making it well-suited for tasks where both modalities are equally important.
    \item \textbf{BLIP} \cite{https://doi.org/10.48550/arxiv.2201.12086}: We utilize this model to bridge the gap between vision and language tasks by effectively leveraging image-language pre-training.
    \item \textbf{ALBEF} \cite{https://doi.org/10.48550/arxiv.2107.07651}: The Alignment of Language and Vision using BERT leverages a dual-transformer structure that synchronizes learning between visual and textual representations.
\end{enumerate} 
\subsection{LLM Based Models. } We used ChatGPT \cite{ouyang2022training}, LLaMA \cite{touvron2023llama}, GPT 4 \cite{openai2024gpt4}, along with Mistral LLMs for zero-shot simple prompt-based and CoT-based models. 
\subsection{Experimental Details}\label{app:im_det}
All models, including baselines, were implemented using the Huggingface Transformers library\footnote{\url{https://huggingface.co/docs/transformers/index}}, 
with a fixed random seed of $42$ for consistency. The details of hyper-parameters are given in the Appendix Table \ref{tab:hyper}. The training was conducted on a single NVIDIA-GTX-1080Ti GPU with 16-bit mixed precision. For the proposed model, hyperparameters $\alpha$, $\beta$, $\gamma$, and $\theta$ in the overall loss function $\mathcal{L}_{\text{F}}$ (Equation \ref{eq:overall}) were determined through grid search and set to 0.5, 0.5, 0.3, and 0.4, respectively.\\
\begin{table}[!htp]\centering
\scriptsize
\adjustbox{width=\linewidth}{\begin{tabular}{lrrrrrr}\toprule
\textbf{Hyper-Parameter} &\textbf{MAMI} &\textbf{MIMIC} &\textbf{Hateful} &\textbf{Memotion2} &\textbf{Harmful} \\\cmidrule{1-6}
epoch &60 &60 &60 &60 &60 \\
batch size &64 &64 &64 &64 &64 \\
Learning Rate &3e-5 &3e-5 &1e-4 &3e-5 &5e-4 \\
Optimizer &Adam &Adam &Adam &Adam &Adam \\
Image Size &224 &224 &224 &224 &224 \\
Random seed &42 &42 &42 &42 &42 \\
\bottomrule
\end{tabular}}
\caption{Details of Hyper-parameters}\label{tab:hyper}
\end{table}
\noindent
\textbf{LLM:} 
For our proposed model M3Hop-CoT, we used Mistral-7B-Instruct-v0.1 \cite{jiang2023mistral} LLM, which has 7 billion parameters. \\

\noindent
\textbf{Tokenizer:} To extract the textual and visual features, we have utilized a pre-trained CLIP (Contrastive Language-Image Pretraining) model. CLIP is a transformer-based architecture focusing solely on the encoder (no decoder) and utilizes contrastive learning to make textual and visual features semantically similar. Our model leverages the CLIP tokenizer, which employs byte pair encoding (BPE) with a lowercase vocabulary of 49,152 tokens. To facilitate model processing, text sequences are padded with special tokens: "[SOS]" at the beginning and "[EOS]" at the end, signifying the start and end of the sequence, respectively.\\

\noindent
For the MAMI dataset, we used the clip (clip-ViT-B-32) model, and for the MIMIC dataset, we utilized multilingual CLIP (mCLIP) (M-CLIP/XLM-Roberta-Large-Vit-L-14).
\begin{figure*}[t!]
\centering
\includegraphics[width=\linewidth]{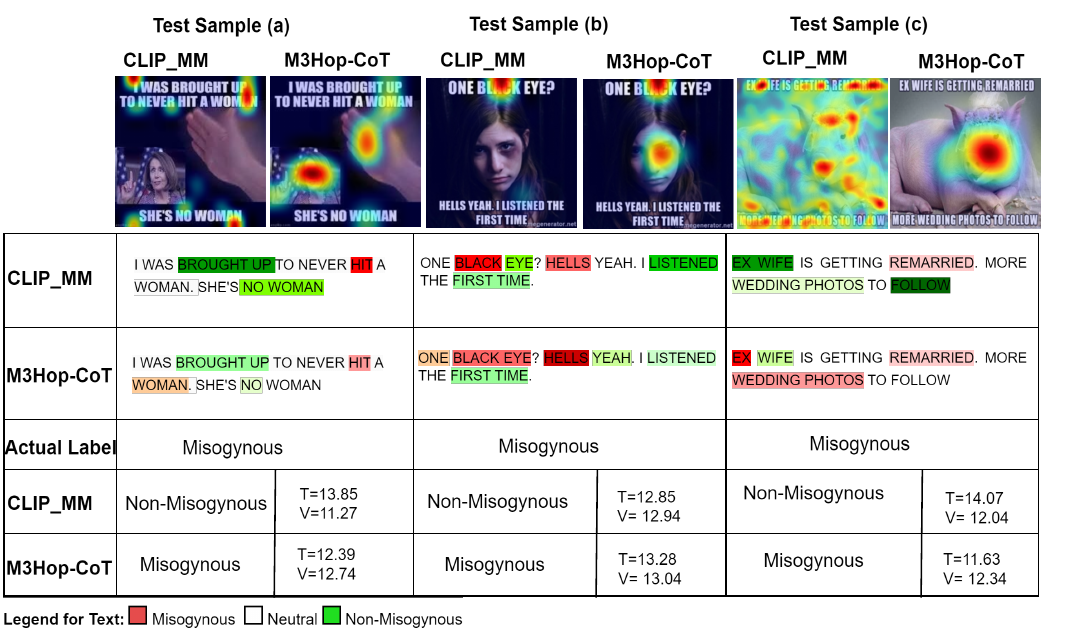}
\caption{Case studies comparing the attention-maps for the baseline \textbf{CLIP\_MM} and the proposed model \textbf{M3Hop-CoT} using Grad-CAM, LIME \cite{DBLP:journals/corr/RibeiroSG16}, and Integrated Gradient \cite{DBLP:journals/corr/SundararajanTY17} on the \textbf{MAMI dataset} test samples. Here, T and V are the normalized textual and visual contribution scores in the final prediction using Integrated Gradient.\label{fig:detailed}}
\end{figure*}

\begin{figure*}[t!]
\centering
\includegraphics[width=\linewidth]{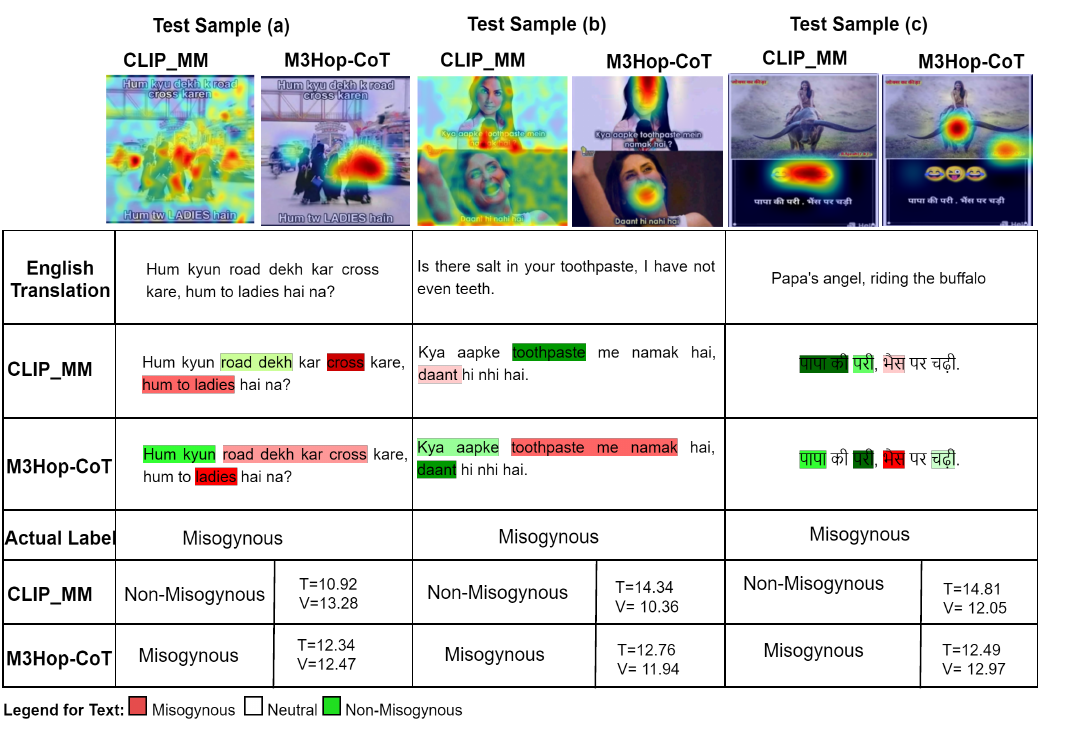}
\caption{Case studies comparing the attention-maps for the baseline \textbf{CLIP\_MM} and the proposed model \textbf{M3Hop-CoT} using Grad-CAM, LIME \cite{DBLP:journals/corr/RibeiroSG16}, and Integrated Gradient \cite{DBLP:journals/corr/SundararajanTY17} on the \textbf{MIMIC dataset} test samples. Here, T and V are the normalized textual and visual contribution scores in the final prediction using Integrated Gradient.\label{fig:detailed_MIMIC}}
\end{figure*}

\begin{figure}[ht!]
    \centering
   \includegraphics[width=\linewidth]{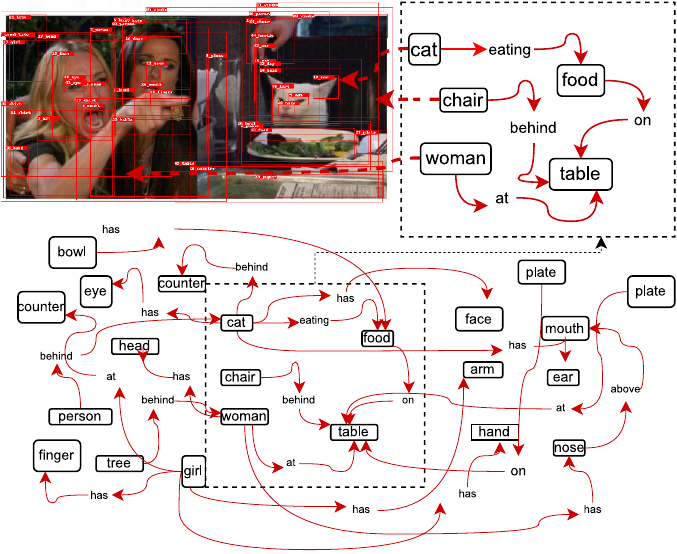}
 \caption{Illustration of scene graph for an image I.\label{fig:scene_graph}}
\end{figure}
\begin{figure}[ht!] 
    \centering
   \includegraphics[width=\linewidth]{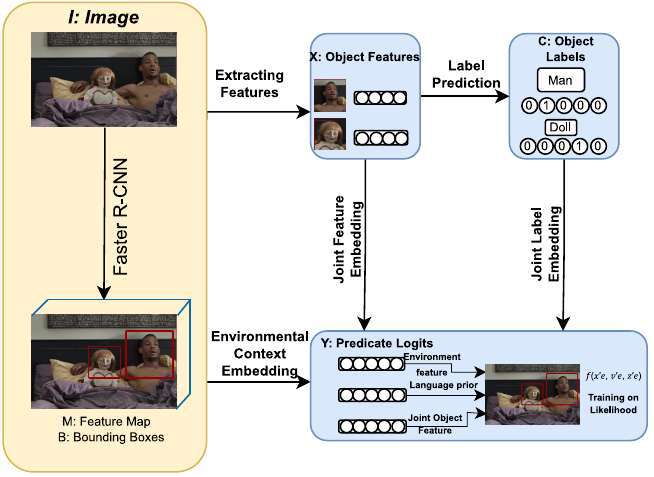}
 \caption{Illustration of the architecture of model used for scene graph.\label{fig:scene_graph_arc}}
\end{figure}

\section{Detailed Discussion of Different LLMs-generated Context and the Impact of Prompts}\label{sec:app_llm}
\begin{figure*}[t!]
    \centering
    \includegraphics[width=\linewidth]{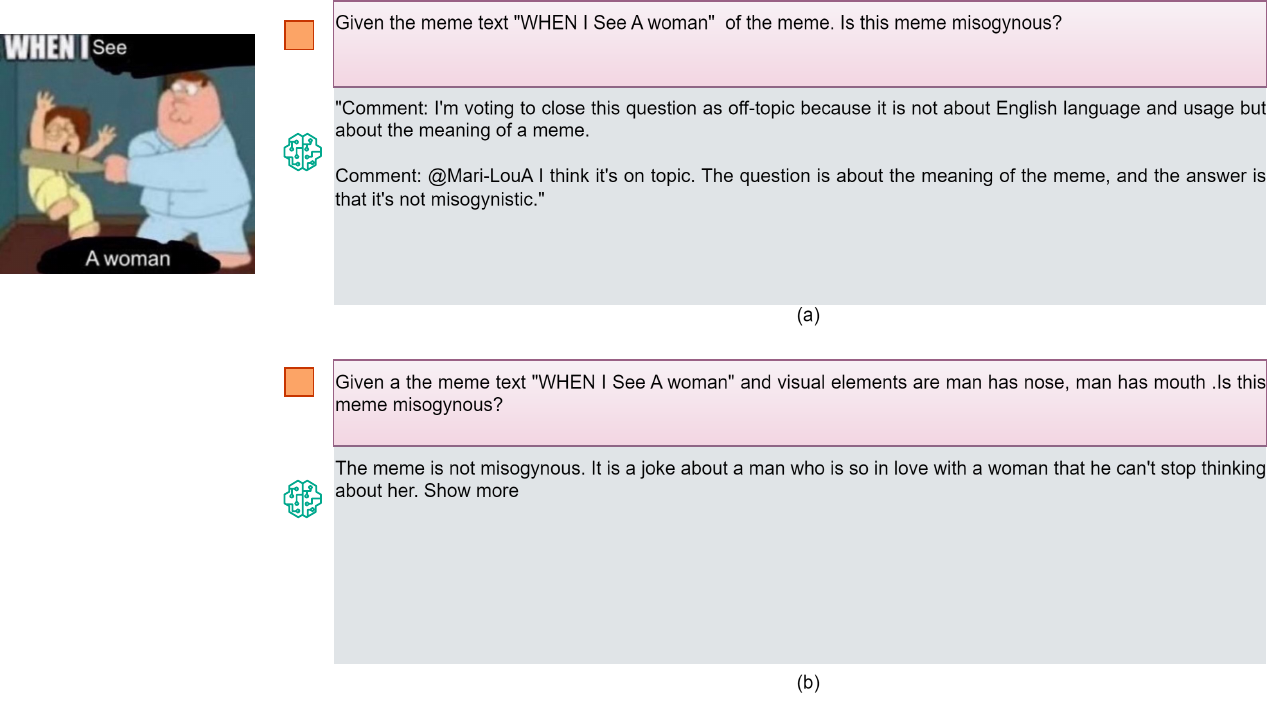}
    \caption{Illustration of context generation using only prompt by Llama LLM for test case 1. (a). Prompt without using Entity-Object-Relationship, (b). Prompt with visual information i.e., using Entity-Object-Relationship \label{fig:cot_prompt_llama_1}}
    \includegraphics[width=\linewidth]{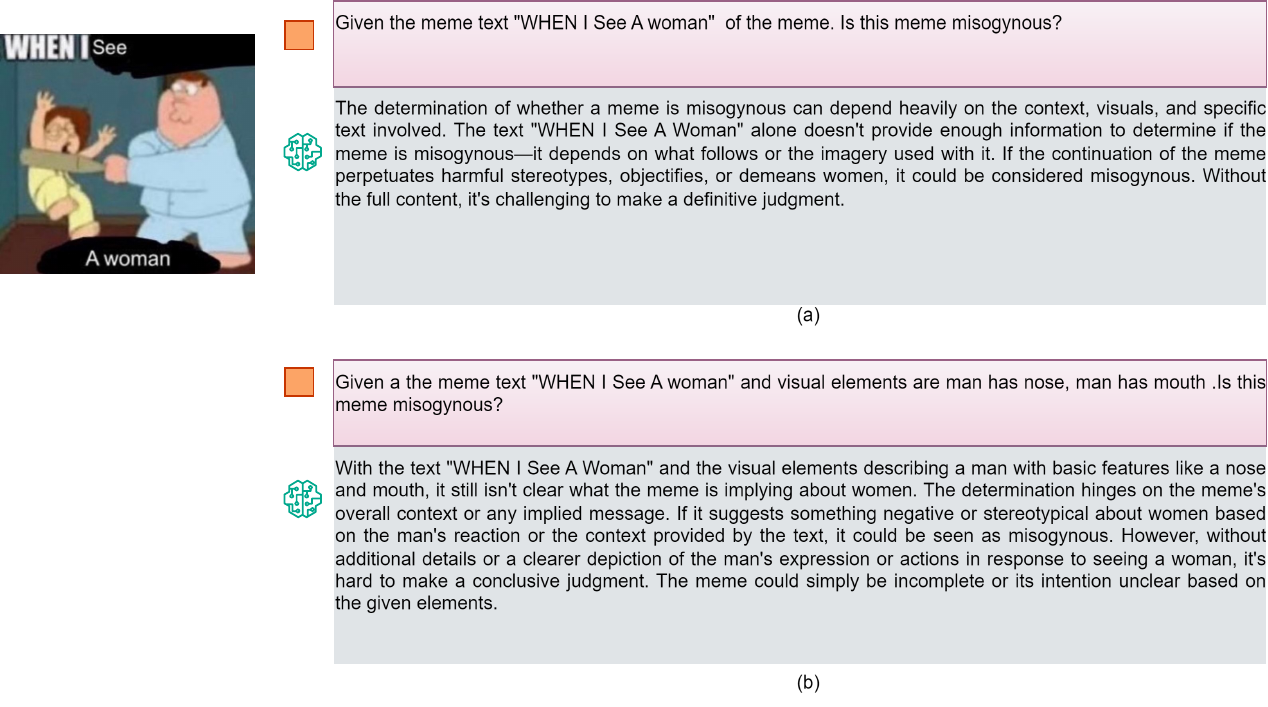}
    \caption{Illustration of context generation using only prompt by ChatGPT LLM for test case 1. (a). Prompt without using Entity-Object-Relationship, (b). Prompt with visual information i.e., using Entity-Object-Relationship \label{fig:cot_prompt_chatgpt_1}}
\end{figure*}

\begin{figure*}[t!]
    \centering
    \includegraphics[width=\linewidth]{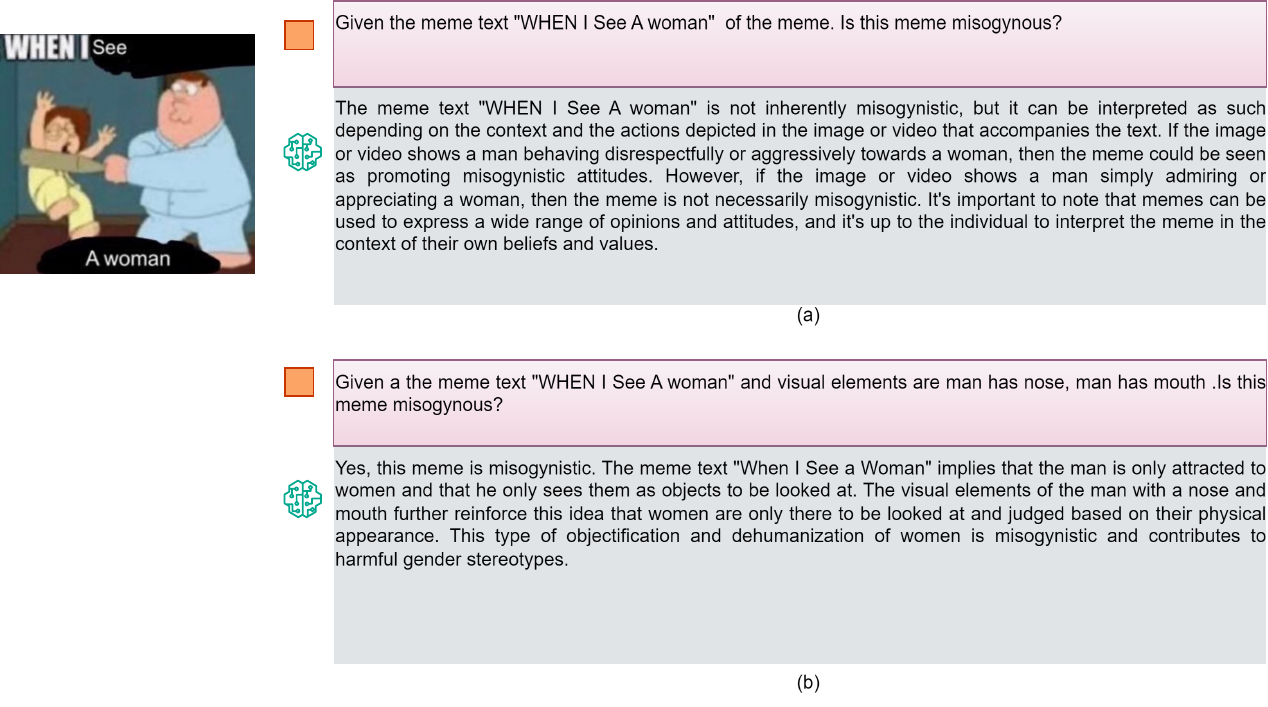}
    \caption{Illustration of context generation using only prompt by Mistral LLM for test case 1. (a). Prompt without using Entity-Object-Relationship, (b). Prompt with visual information i.e., using Entity-Object-Relationship \label{fig:cot_prompt}}
    \includegraphics[width=\linewidth]{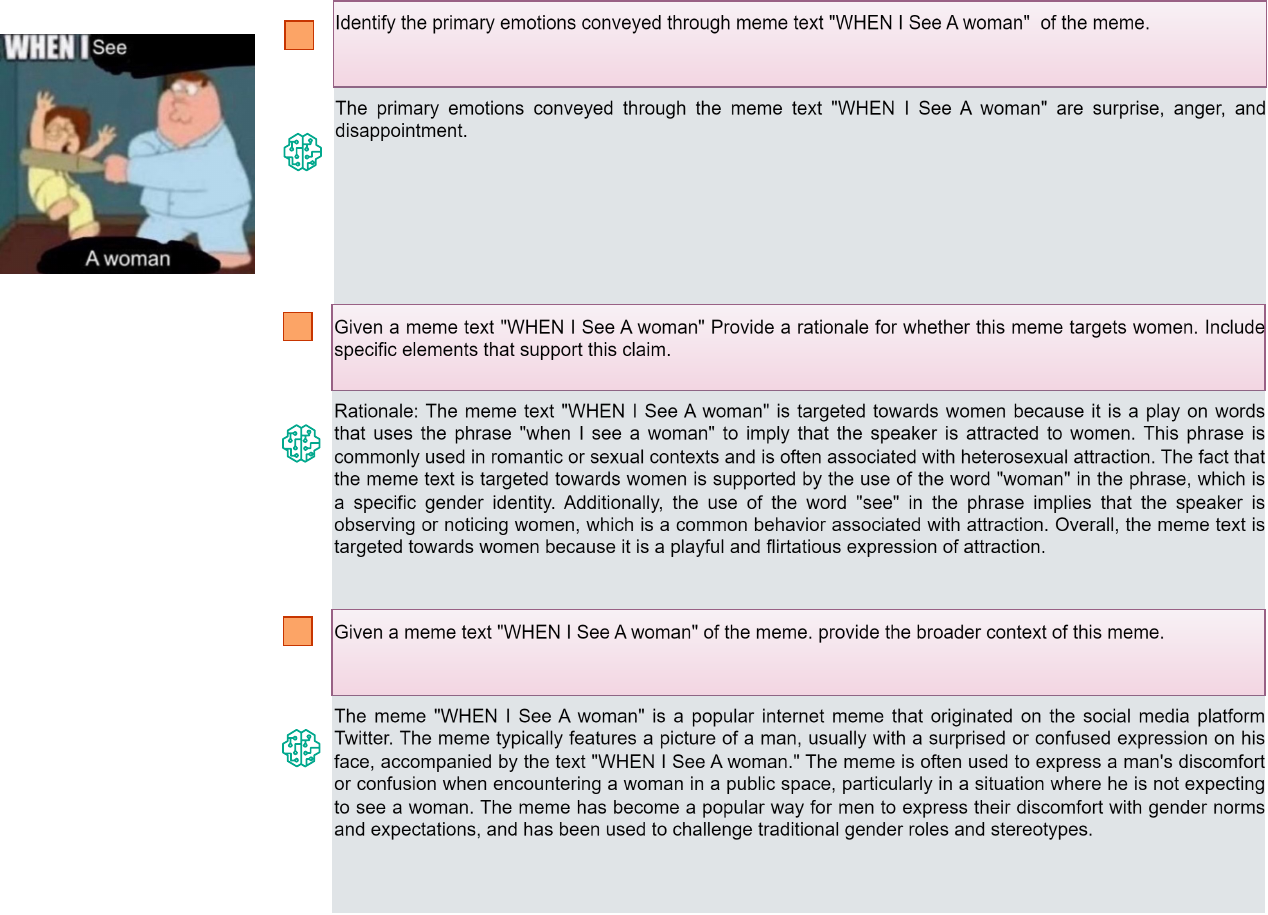}
    \caption{Illustration of context generation  \textcolor{red}{\textbf{without scene graph}} using our CoT prompt with Mistral LLM for test case 1.\label{fig:misteral_cot_prompt}}
    % \includegraphics[width=\linewidth]{ARR_april_LLAMA.pdf}
    % \caption{Illustration of context generation using our CoT prompt with Llama LLM\label{fig:llama_cot_prompt}}
    
\end{figure*}

\begin{figure*}[t!]
    \centering
    \includegraphics[width=\linewidth]{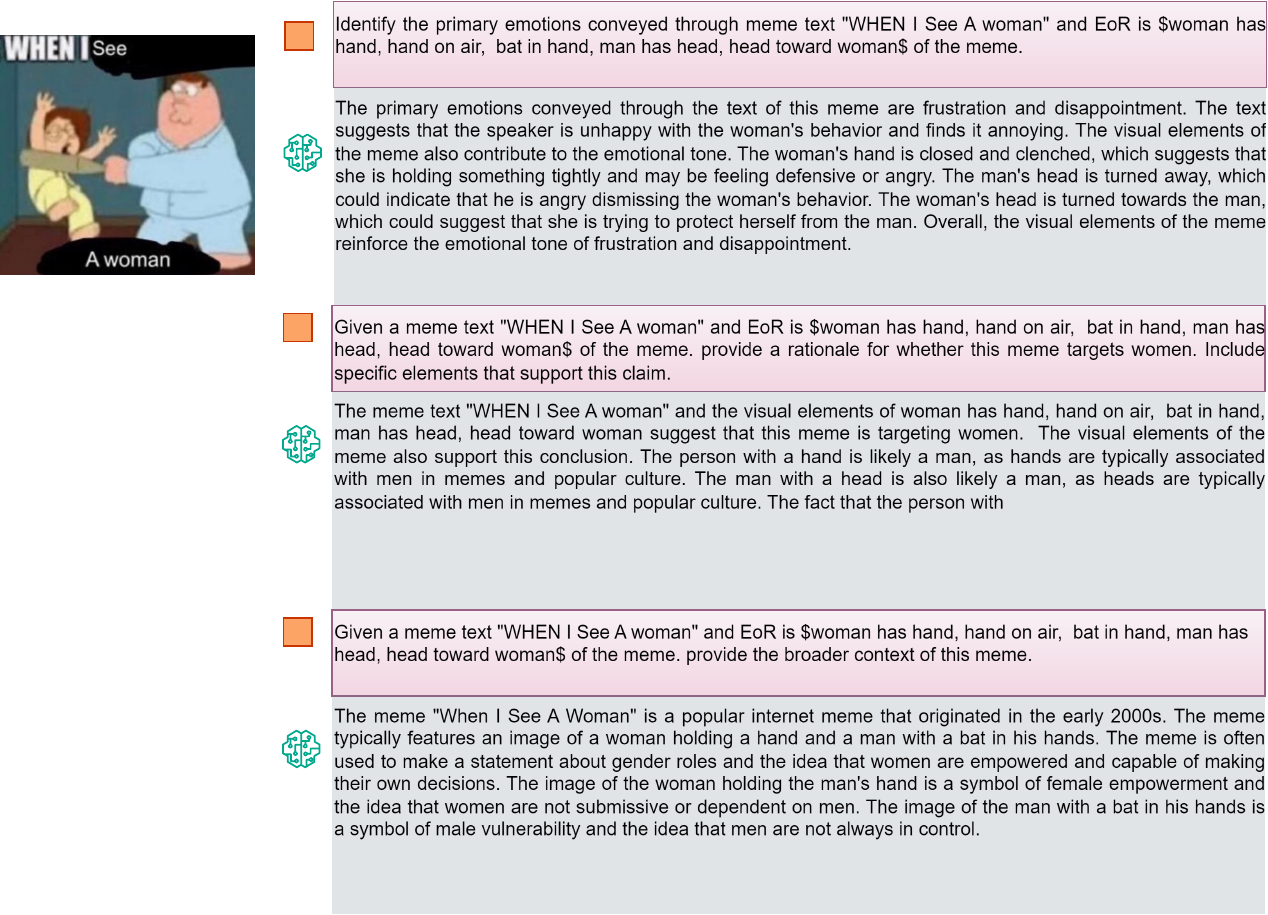}
    \caption{Illustration of context generation using our CoT prompt with Mistral LLM for test case 1.\label{fig:cot_prompt_M}}
    \includegraphics[width=\linewidth]{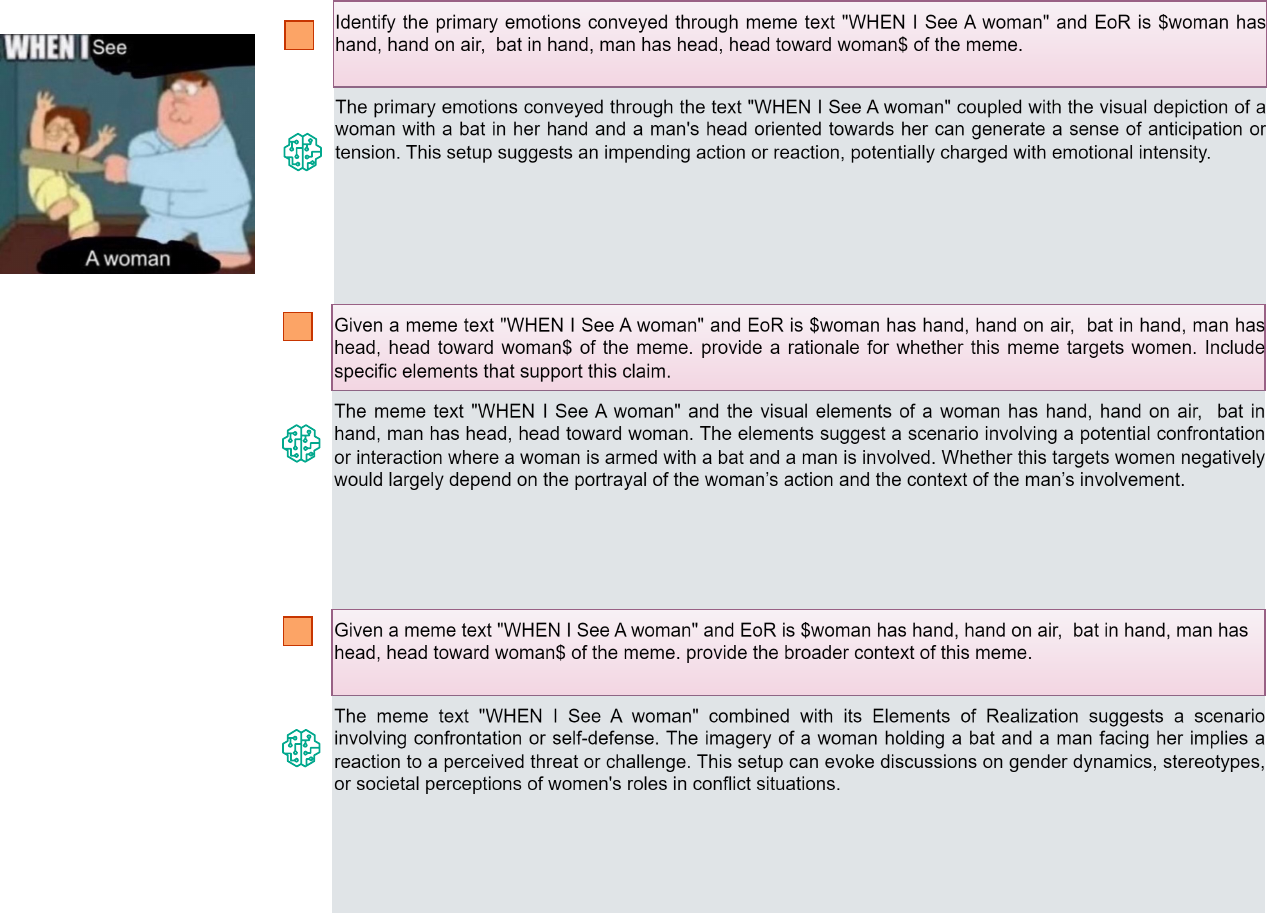}
    \caption{Illustration of context generation using our CoT prompt with ChatGPT LLM for test case 1.\label{fig:chat_gpt_cot_prompt_1}}
    % \includegraphics[width=\linewidth]{ARR_april_LLAMA.pdf}
    % \caption{Illustration of context generation using our CoT prompt with Llama LLM\label{fig:llama_cot_prompt}}
    
\end{figure*}

\begin{figure*}[ht!]
    \centering
   \includegraphics[width=\linewidth]{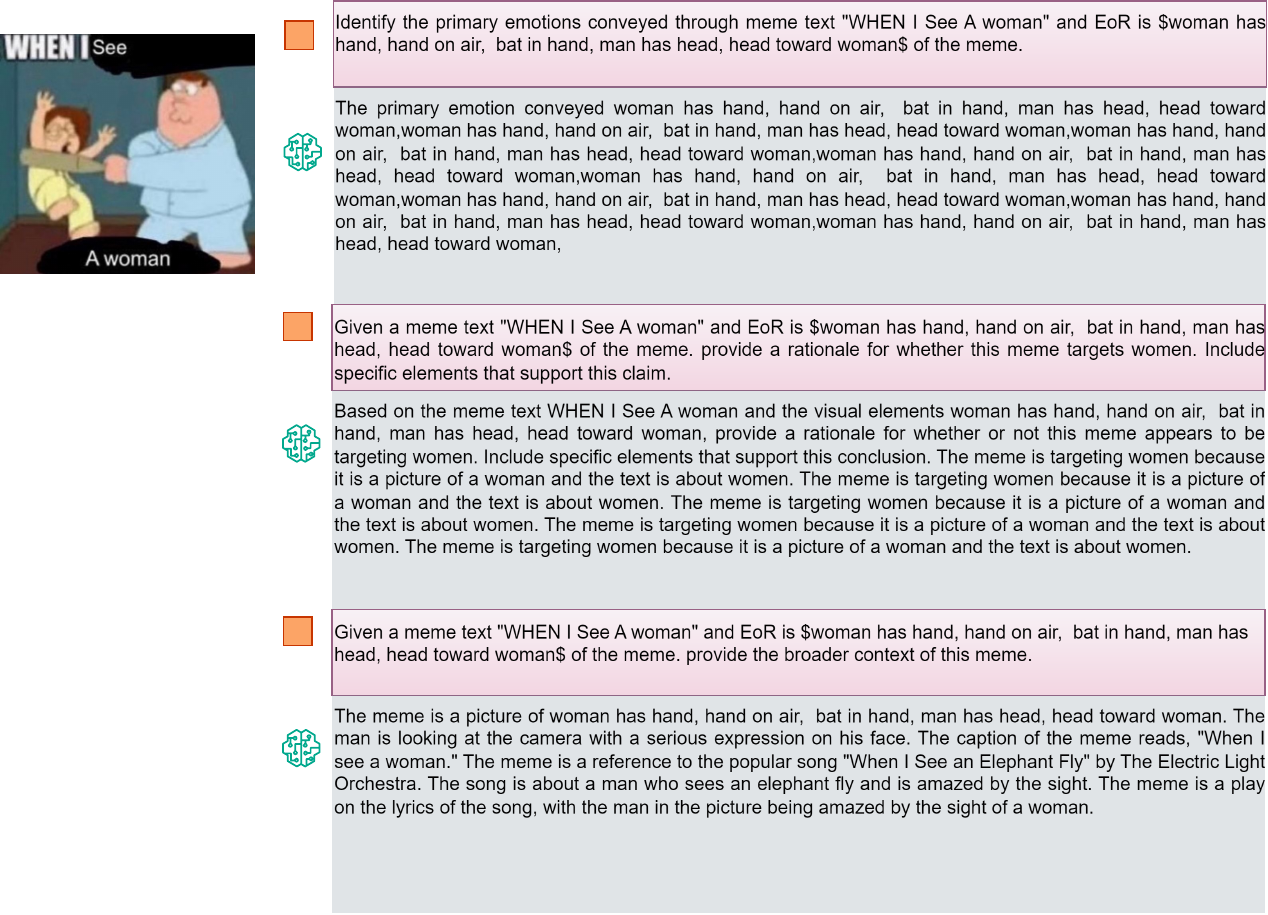}
 \caption{Illustration of context generation using our CoT prompt with Llama LLM for test case 1.\label{fig:llama_cot_prompt}}
\end{figure*}

\begin{figure*}[t!]
    \centering
    \includegraphics[width=\linewidth]{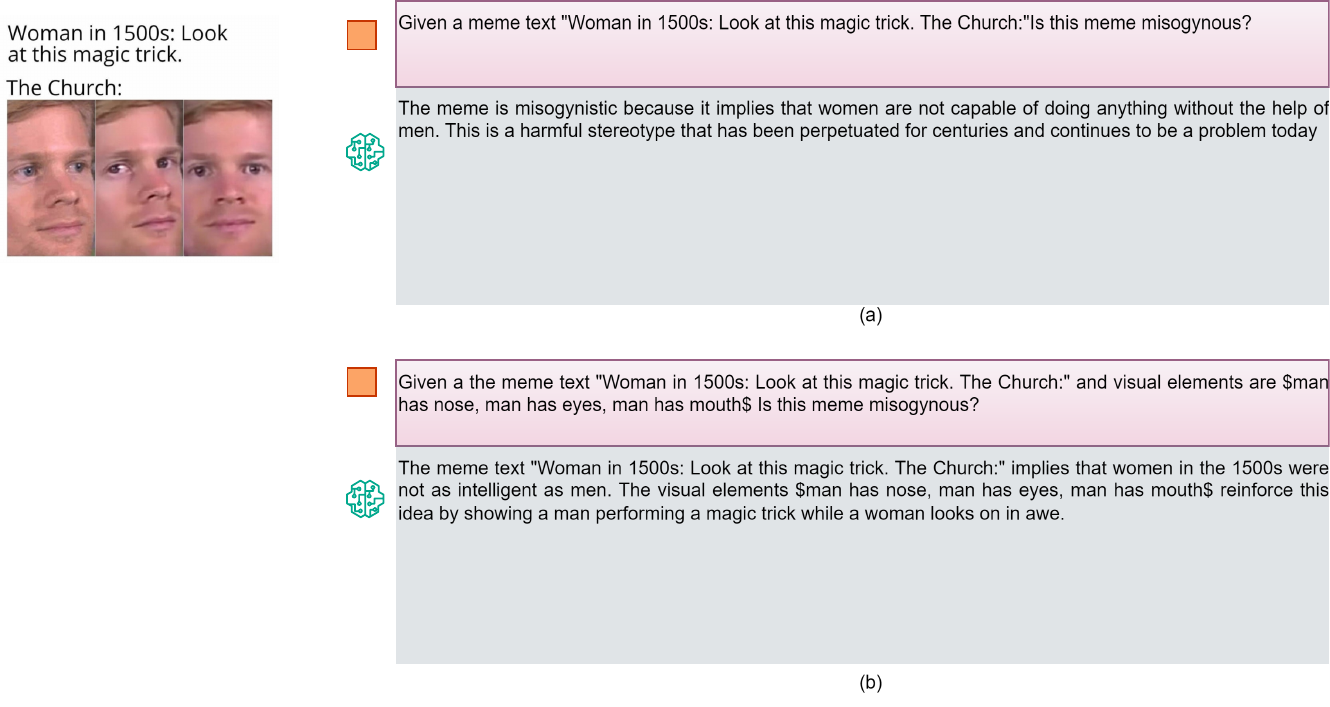}
    \caption{Illustration of context generation using only prompt by Llama LLM for test case 2. (a). Prompt without using Entity-Object-Relationship, (b). Prompt with visual information i.e., using Entity-Object-Relationship \label{fig:prompt_test_case2_llama}}
    \includegraphics[width=\linewidth]{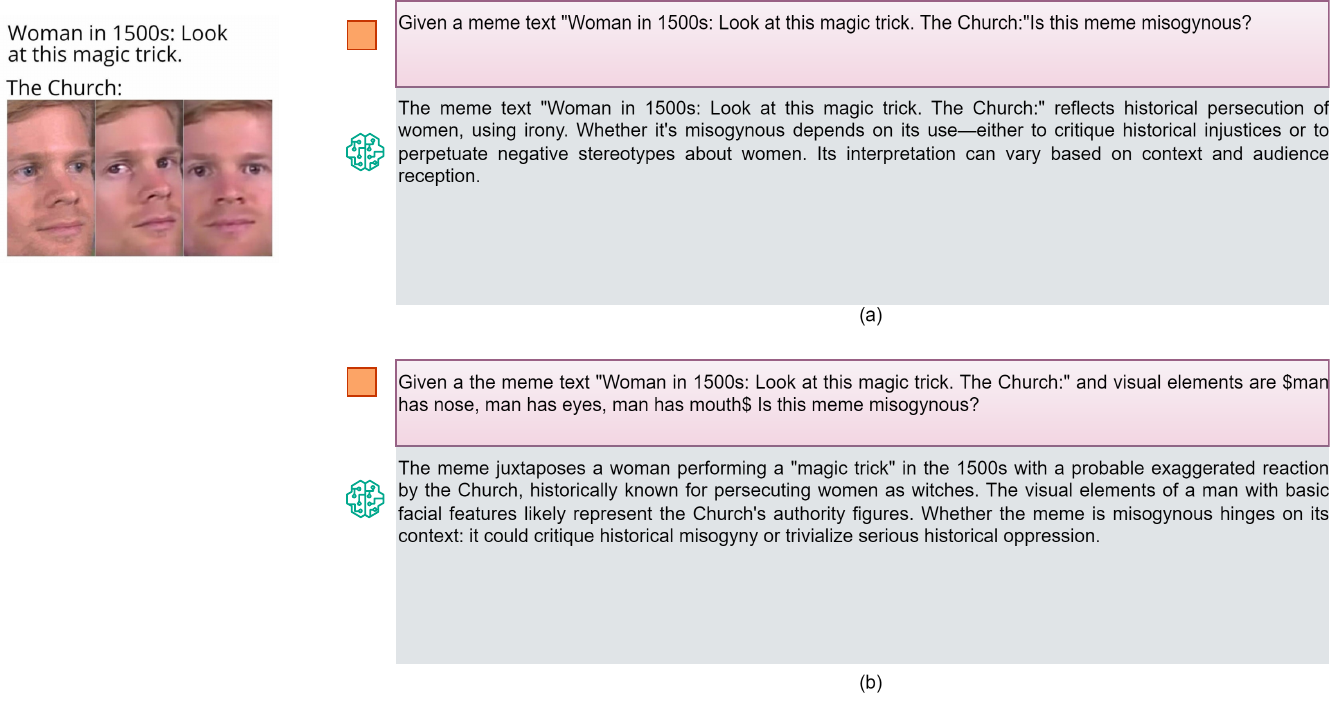}
    \caption{Illustration of context generation using only prompt by ChatGPT LLM for test case 2. (a). Prompt without using Entity-Object-Relationship, (b). Prompt with visual information i.e., using Entity-Object-Relationship \label{fig:prompt_test_case2_chatgpt}}
    % \includegraphics[width=\linewidth]{ARR_april_LLAMA.pdf}
    % \caption{Illustration of context generation using our CoT prompt with Llama LLM\label{fig:llama_cot_prompt}}
    
\end{figure*}

\begin{figure*}[t!]
    \centering
    \includegraphics[width=\linewidth]{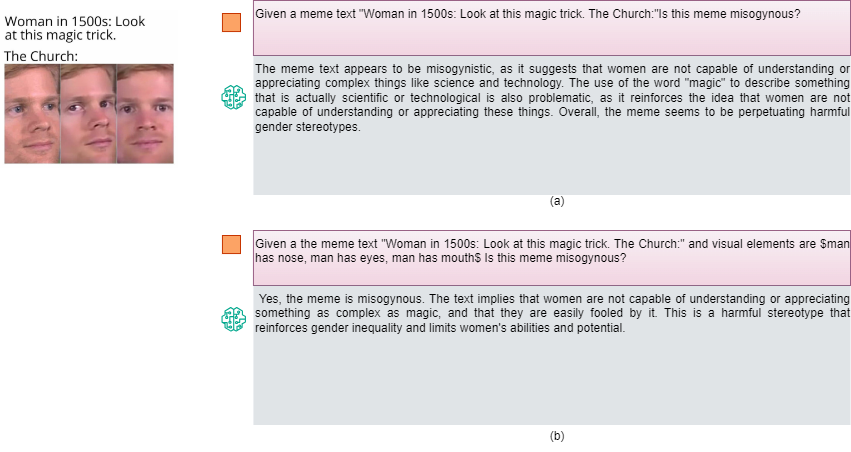}
    \caption{Illustration of context generation using only prompt by Mistral LLM for test case 2. (a). Prompt without using Entity-Object-Relationship, (b). Prompt with visual information i.e., using Entity-Object-Relationship \label{fig:cot_prompt_2}}
    \includegraphics[width=\linewidth]{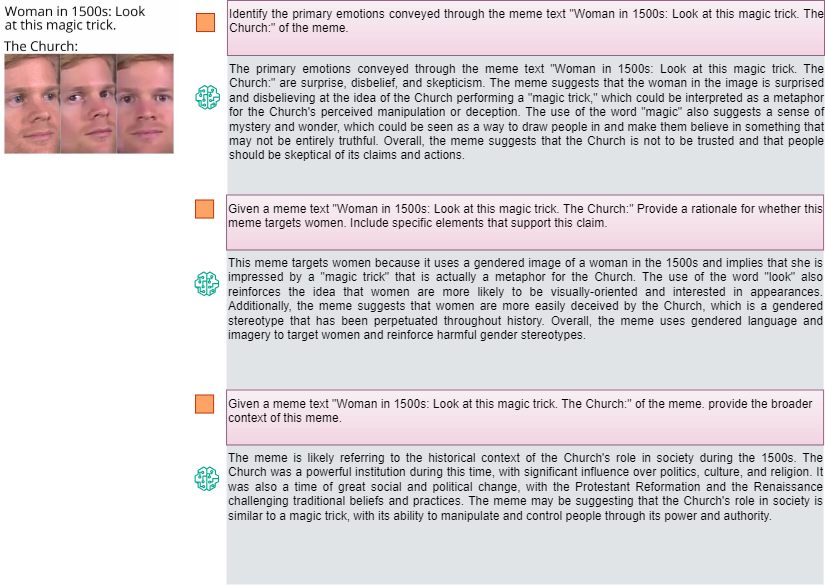}
    \caption{Illustration of context generation  \textcolor{red}{\textbf{without scene graph}} using our CoT prompt with Mistral LLM for test case 2.\label{fig:Mist_cot_prompt_2}}
\end{figure*}

\begin{figure*}[ht!]
    \centering
   \includegraphics[width=\linewidth]{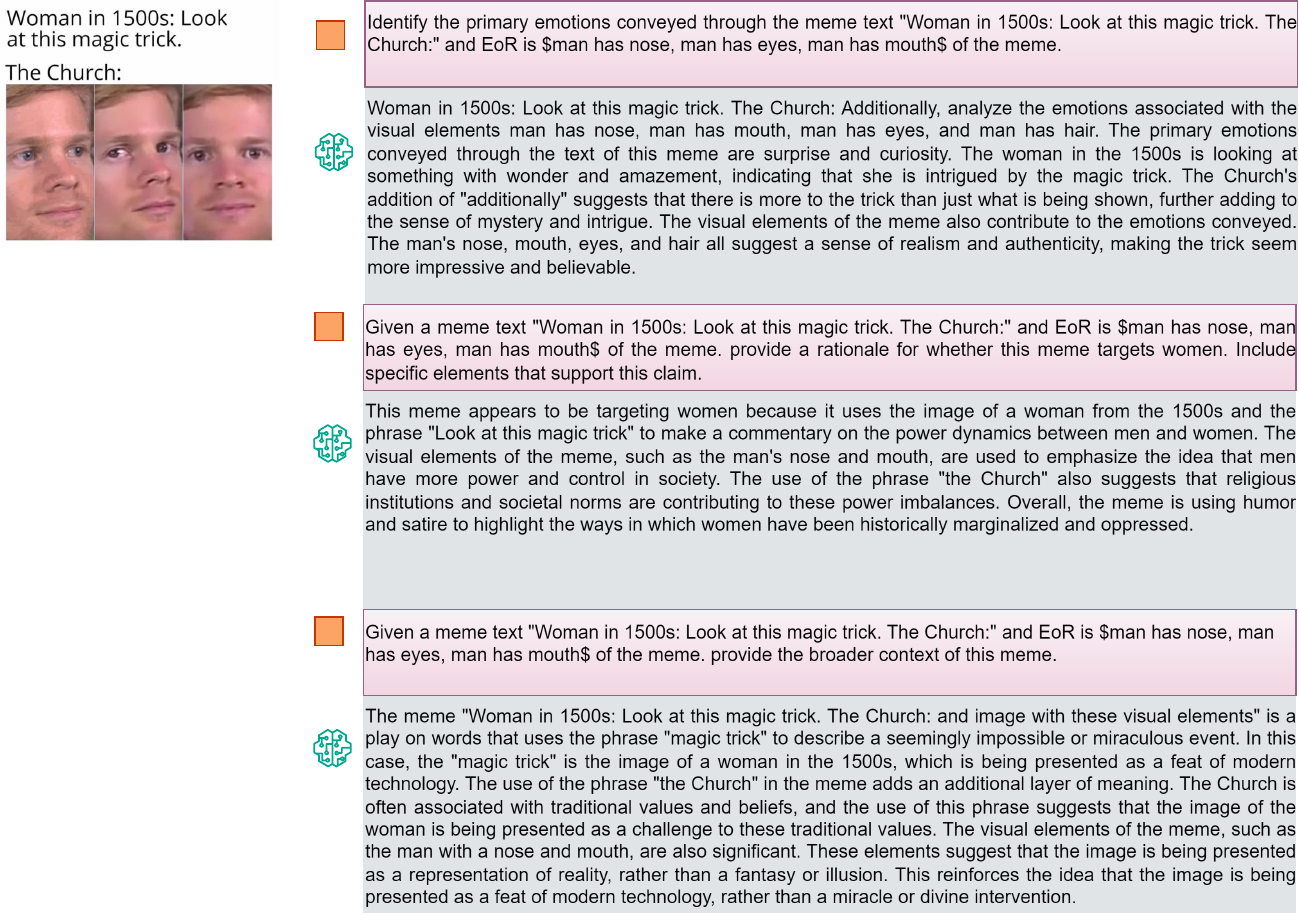}
 \caption{Illustration of context generation using our CoT prompt with Mistral LLM for testing case 2.\label{fig:cot_prompt_ex2}}
\end{figure*}

\begin{figure*}[ht!]
    \centering
   \includegraphics[width=\linewidth]{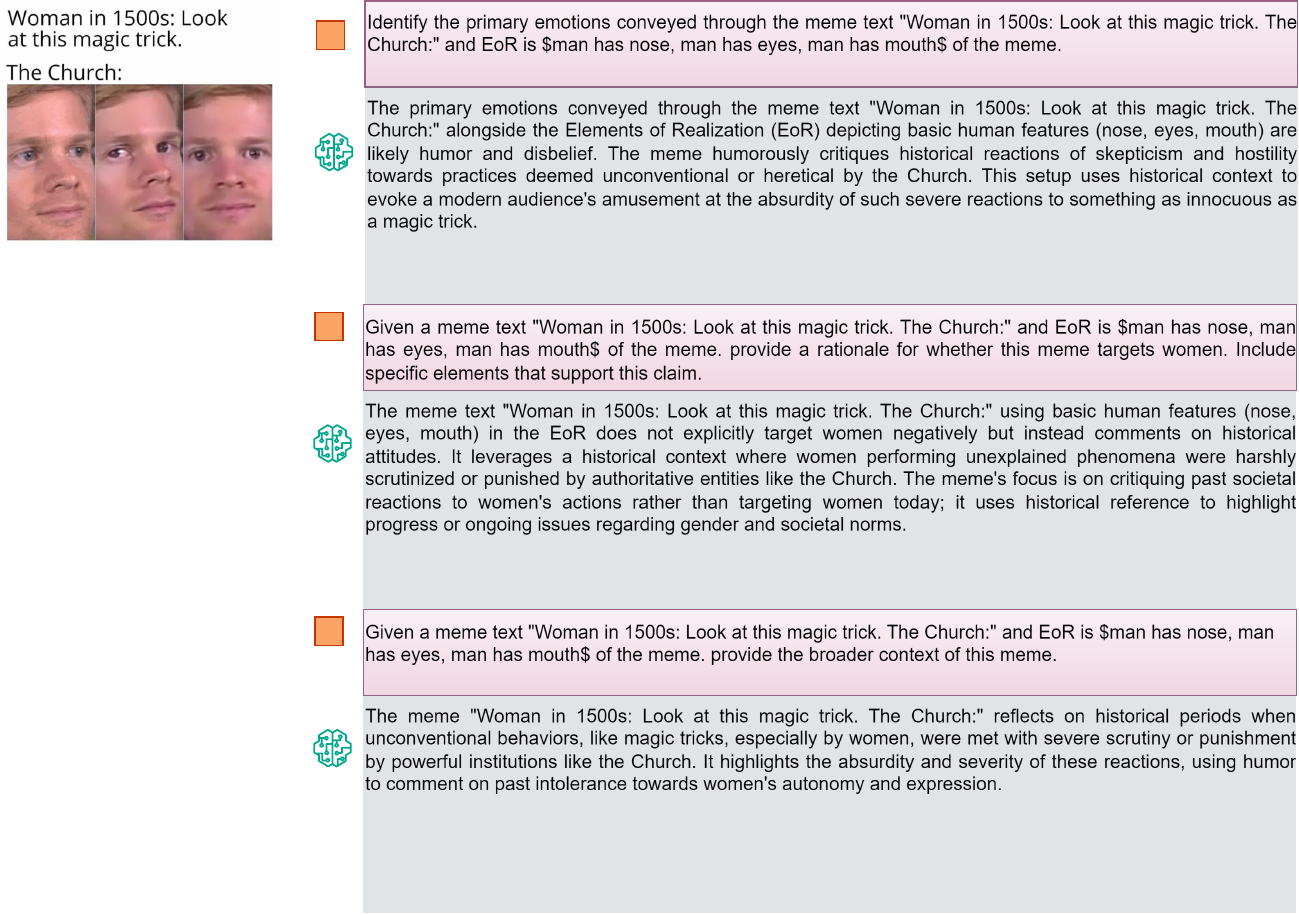}
 \caption{Illustration of context generation using our CoT prompt with ChatGPT LLM for testing case 2\label{fig:chat_gpt_cot_prompt_2}}
\end{figure*}

\begin{figure*}[ht!]
    \centering
   \includegraphics[width=\linewidth]{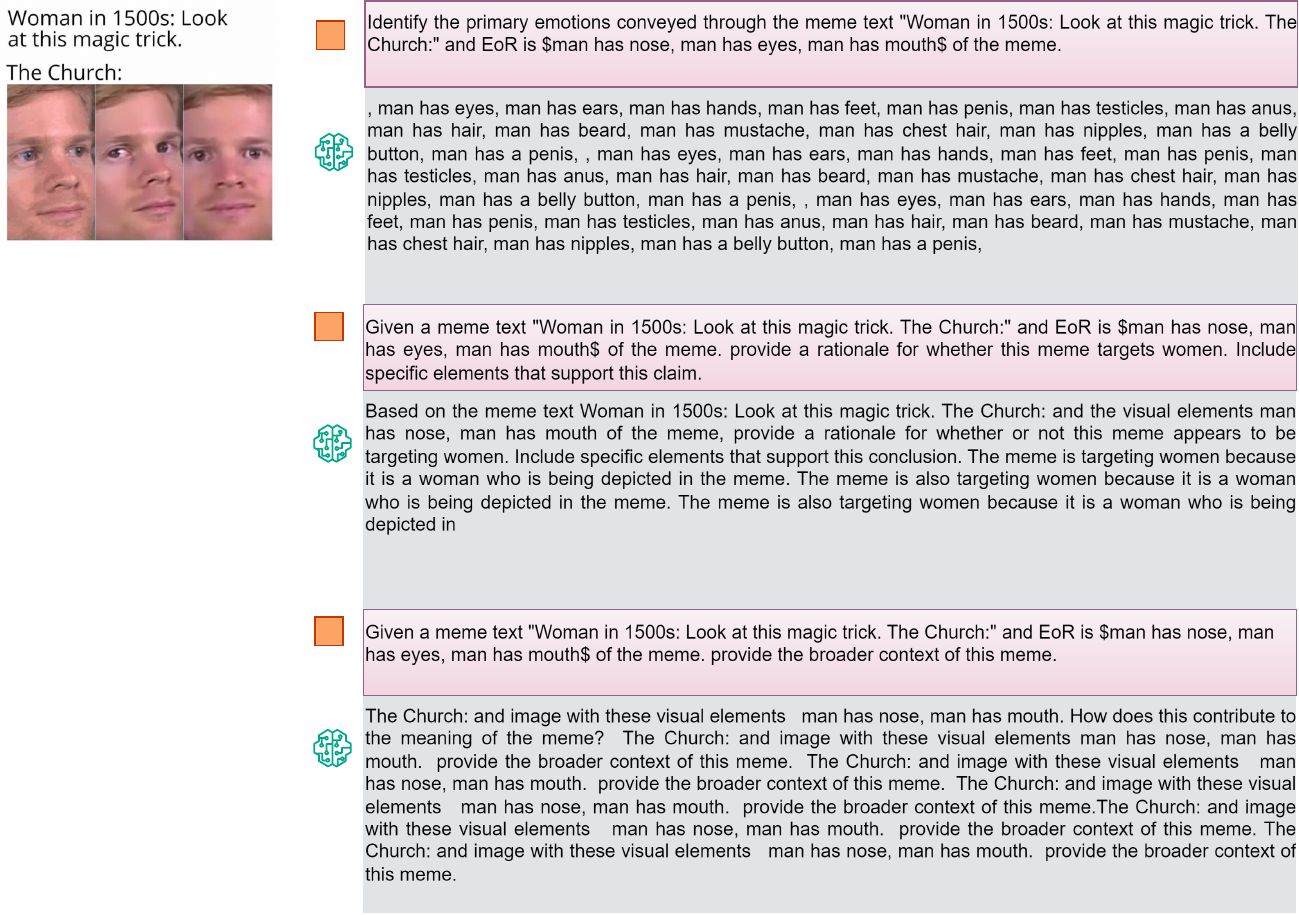}
 \caption{Illustration of context generation using our CoT prompt with Llama LLM for testing case 2.\label{fig:llama_cot_prompt_2}}
\end{figure*}

In the results Table \ref{tab:result}, we have shown the results of four highly robust LLMs ((a) ChatGPT, (b) GPT 4, (c) Llama and (d) Mistral (Ours) ) on the MAMI and MIMIC datasets. We have shown four variations of each LLM used: (i) utilizing simple prompts with only meme text, (ii) utilizing multimodal prompts with meme text with EoRs, (iii) utilizing simple prompts with only meme text by CoT technique, and (iv) utilizing multimodal prompts with meme text with EoRs by CoT technique.
\begin{enumerate}[nolistsep,noitemsep]
    \item \textbf{Utilizing only meme text as prompt to LLM: } The first example presents a meme with the text "When I see a woman," accompanied by an image depicting violence against a woman ( Refer to Figures \ref{fig:cot_prompt_llama_1} (a), \ref{fig:cot_prompt_chatgpt_1}(a), \ref{fig:cot_prompt} (a)). In the absence of the image modality by EOR, all three LLMs (ChatGPT, Llama, and Mistral) struggle to identify the misogynistic content within the meme. Llama and ChatGPT respond with uncertainty, indicating a dependence on the broader context. Conversely, Mistral offers a distinct response, stating that the text itself is not inherently misogynistic but could be interpreted as such depending on accompanying visual information. This suggests that Mistral, unlike the other models, attempts to generate context similar to humans even without additional modalities. This behavior presents a promising avenue for further exploration and development. The second example (Figure \ref{fig:prompt_test_case2_llama} (a), \ref{fig:prompt_test_case2_chatgpt} (a), \ref{fig:cot_prompt_2} (a)) presents a meme with the text "Woman in 1500s: Look at the magic trick. The Church." This meme appears to mock and portray a woman's alleged inability to comprehend 15th-century technology, juxtaposed with the Church exhibiting a similar lack of understanding. Here also, unlike other LLMs, Mistral provides a comprehensive explanation, elucidating how this meme perpetuates stereotypes targeting women. Mistral highlights the meme's role in disseminating and reinforcing gender-based and negative stereotypes about the Church.
    \item \textbf{Utilizing multimodal prompts with meme text and EoRs to LLM:} Now, comparing only text-based LLMs with multimodal LLMs by adding EORs surely adds the extra benefit of the model's understanding of memes while making the prediction. In Figures \ref{fig:cot_prompt_llama_1}(b), \ref{fig:cot_prompt_chatgpt_1}(b), and \ref{fig:cot_prompt}(b), the prompt "When I see a woman" is presented alongside an image of man depicting violence against a woman. Including visual elements as EoRs significantly enhances understanding of the meme's context for all LLMs. The Figures \ref{fig:prompt_test_case2_llama} (b), \ref{fig:prompt_test_case2_chatgpt} (b), \ref{fig:cot_prompt_2} (b), demonstrate that EoRs alone are insufficient for grasping the deeper meaning of memes with implicit offensiveness, as seen in the meme text "Woman in 1500s: Look at the magic trick. The Church:". This highlights the limitations of EoRs and underscores the need for a CoT-based approach, as employed by our proposed model, for comprehensive understanding.
    \item \textbf{Utilizing only meme text as prompt by CoT technique:} Now, we have explored the effectiveness of CoT prompting using only meme text to understand the need for human-like reasoning for identifying misogyny. Figures \ref{fig:misteral_cot_prompt} and \ref{fig:Mist_cot_prompt_2} showcase the rationale generated by the Mistral model using the CoT technique. This rationale analyzes emotions, targets, and context within the meme. As seen in Figure \ref{fig:misteral_cot_prompt}, the model identifies emotions like surprise, anger, and disappointment, along with the targeted nature and context of the meme. This analysis helps our proposed model (M3Hop-CoT) to understand the underlying claim of degrading women and make a correct prediction. Similarly, Figure \ref{fig:Mist_cot_prompt_2} demonstrates how human-like reasoning, achieved through CoT prompting, allows M3Hop-CoT to analyze the three crucial cues (emotions, target, and context) and accurately identify the misogynistic label. However, without a visual element, the LLM fails to generate accurate reasoning about the meme.
    \item \textbf{Utilizing multimodal prompts with meme text and EoRs by CoT technique:} Finally, we showcase the impact of CoT prompting with multimodal prompts, incorporating both meme text EoRs. Figures \ref{fig:cot_prompt_M}, \ref{fig:chat_gpt_cot_prompt_1}, \ref{fig:llama_cot_prompt}, and corresponding Figures \ref{fig:cot_prompt_ex2}, \ref{fig:chat_gpt_cot_prompt_2}, and \ref{fig:llama_cot_prompt_2} present the rationale generated by different LLMs using the CoT technique for the same two examples. The results demonstrate the superiority of Mistral with multimodal prompts. As seen in Figures \ref{fig:cot_prompt_M} and \ref{fig:cot_prompt_ex2}, Mistral generates highly relatable, human-like rationales for both examples. These rationales consider emotions, targets, context, and cultural nuances within the meme. Compared to Mistral, LLMs like ChatGPT and Llama struggle to produce such comprehensive and culturally rich rationales (Figures \ref{fig:chat_gpt_cot_prompt_1}, \ref{fig:chat_gpt_cot_prompt_2}, \ref{fig:llama_cot_prompt}, \ref{fig:llama_cot_prompt_2}). This highlights the effectiveness of Mistral in leveraging CoT prompting with multimodal information for superior performance in misogyny detection.
\end{enumerate}

\section{Result analysis on Hateful meme, Memotion2 and Harmful meme dataset}\label{sec:detail_generic_data}
To evaluate the robustness of our proposed method across various datasets and to understand how common, language-specific taboo elements affect generalization, we conducted a comprehensive generalization study, as highlighted in \citep{nozza-2021-exposing,10.1145/3457610,ranasinghe-zampieri-2020-multilingual}. We tested our model on three well-known datasets: the Hateful Memes dataset \citep{NEURIPS2020_1b84c4ce}, the Memotion dataset \citep{sharma-etal-2020-semeval}, and the Harmful Memes dataset. Notably, these datasets are predominantly in English and were used to evaluate our model in a zero-shot manner, meaning the model was not directly trained on these specific datasets. These datasets include unique linguistic elements, such as slang and jargon that differ significantly from those found in misogynous memes, making them challenging out-of-distribution samples for our model.\\

\noindent
Despite these challenges, our model demonstrated robust performance across all datasets, underscoring its effectiveness as shown in Table \ref{tab:result_gen} and illustrating its broad applicability. The model's ability to handle linguistic and cultural variances effectively showcases its versatility and potential for widespread use across diverse data sources. The performance metrics from the Hateful Memes dataset, detailed in Table \ref{tab:result_gen}, offer valuable insights into how our novel M3Hop-CoT model compares with various baseline and state-of-the-art (SOTA) models.

\subsection{Results on Memotion meme dataset}
Our proposed model's performance through zero-shot learning on the Memotion dataset gives balanced results. While pre-trained vision and language models deliver significantly good results, the prompt-based model, PromptHate, surpasses all other models, highlighting the efficacy of prompting techniques. However, when evaluating the performance of our proposed model, it shows improvement over other pre-trained models, but the increment is not significant. The performance increment is 5.03\% lower compared to the Hateful Meme dataset. This discrepancy can be attributed to the different nature of the Memotion dataset. Unlike the Hateful Memes dataset, which is synthetically generated following a specific template, the Memotion dataset comprises real data collected from social media platforms and features a variety of generalized templates. Considering the real-world nature of the Memotion dataset, an improvement of 5.03\% is nevertheless substantial, demonstrating the transferability and robustness of our M3Hop-CoT model. This supports our hypothesis that our model effectively captures critical aspects such as emotion, target, and context of the memes, consistent with the observed trend where social media content frequently targets women. This outcome underscores the prevalence and impact of gender-targeted content in the social media discourse.\\

Now, when we trained our model on the entire dataset using simple prompts and CoT prompts with LLMs, we obtained better results than the PromptHate, Momenta, and even the SOTA model ($\backsim6$\% increment). It shows the efficiency of LLMs' understanding of meme's hidden emotions, targeted knowledge, and contextual information, which helped the model outperform baselines.\\

\begin{figure*}[ht!]
    \centering
    \includegraphics[width=\linewidth]{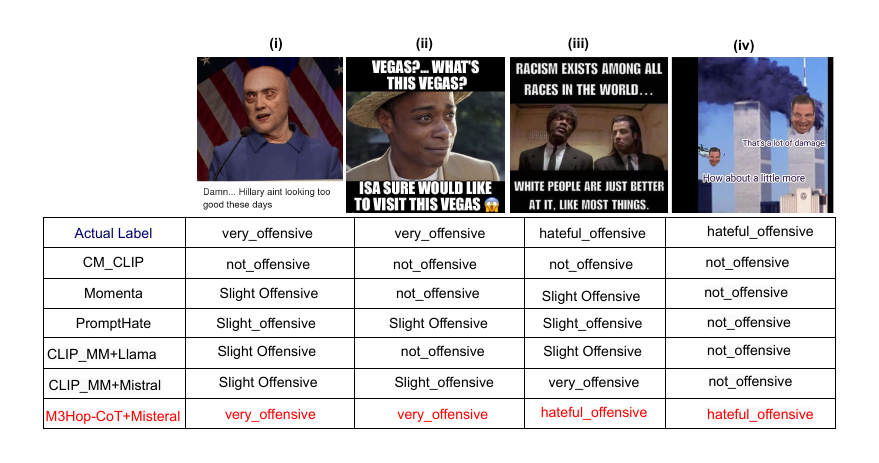}
 \caption{Predictions from different models on a few test samples from Memotion Dataset.\label{fig:memotion_meme}}
\end{figure*}

To illustrate the effectiveness of our proposed model, M3Hop-CoT, over baseline and SOTA models, we present a few examples from the memotion dataset in Figure \ref{fig:memotion_meme}. Each example highlights the crucial role emotions, contextual information, and targeted knowledge play in accurately identifying meme labels. For instance, sample (i) shows an offensive image degrading a political leader. While baseline and SOTA models fail to capture these cues, M3Hop-CoT leverages its LLM strength to provide rationales based on three key elements: emotions, context, and targeted knowledge. Similarly, in samples (ii) and (iii), where indirect racism is subtly conveyed within the memes, M3Hop-CoT accurately identifies the offensive nature of the memes. These findings demonstrate M3Hop-CoT's enhanced ability to understand the nuances of complex memes compared to existing models.
\subsection{Results on Hateful meme dataset}
The Hateful Memes dataset is specifically designed for a hateful meme challenge by synthetically generating memes that alter keywords and images within a template context. Despite being synthetically generated, the Hateful Memes dataset is a task-specific dataset, similar to the MAMI dataset. It encompasses a broad spectrum of what constitutes hate, which also explicitly includes samples that are offensive towards women. This allows for targeted analysis of hate speech and misogyny within memes, reflecting the complexity of the issues being addressed.
\begin{figure*}[ht!]
    \centering
   \includegraphics[width=\linewidth]{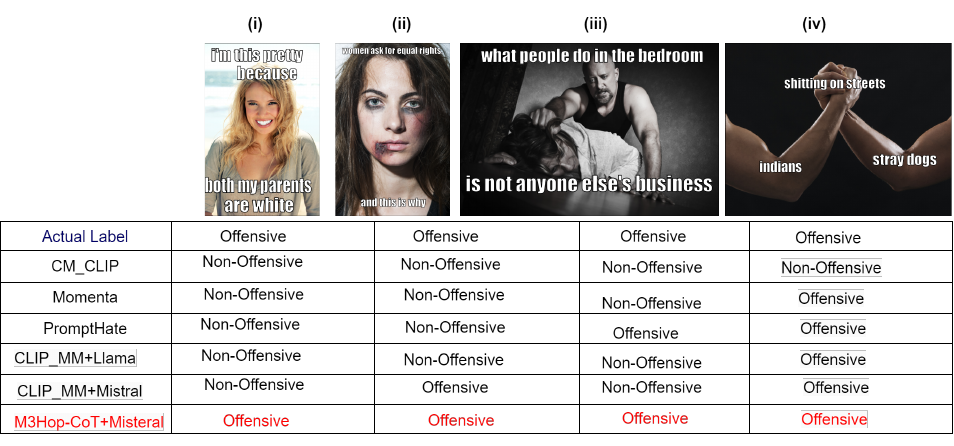}
 \caption{Predictions from different models on a few test samples from hateful meme dataset.\label{fig:hateful_meme}}
\end{figure*}

Consequently, when evaluating the Hateful Memes dataset using our M3Hop-CoT model, we achieved remarkably consistent results. Our analysis showed that pre-trained models like VisualBERT, CLIP, and ALBEF also performed well. Among prompt-based models, PromptHate exhibited high accuracy on this dataset. However, a deeper analysis highlighted the significance of our M3Hop-CoT model, which extends beyond mere prompting techniques. M3Hop-CoT enhances its capability by integrating cultural diversity through hierarchical prompts and effectively utilizing Entity-Object Relationships (EoRs), providing a more nuanced understanding of the memes.\\

Now, training M3Hop-CoT on the complete dataset yielded superior performance compared to all baseline and state-of-the-art (SOTA) models, as shown in Table \ref{tab:result_gen}. For a qualitative comparison, in Figure \ref{fig:hateful_meme}, we have shown four meme samples with the actual label of ``Offensive." where offensiveness in all the memes are implicit in nature. Compared to the SOTA models like Momenta and PromptHate, M3Hop-CoT excels at identifying implicit offensiveness. While these models leverage image entities and captions respectively, still they fail to grasp the underlying meaning. Notably, LLMs (CLIP\_MM+Mistral) are helping in recognizing subtle cues within samples (ii) and (iv). However, LLMs alone struggle with more complex patterns of offensiveness, as seen in samples (i) and (iii). In such cases, M3Hop-CoT's with CoT-based prompting approach, mimicking human reasoning, empowers it to predict the offensive label accurately.

\subsection{Results on Harmful meme dataset}
When evaluating our M3Hop-CoT model using the Harmful Memes dataset, we observed a greater difference in the results compared to the MAMI and other datasets. The Harmful Memes predominantly focuses on the domain of U.S. politics and COVID-19, and both the textual and visual modality of these memes differ significantly from those in the MAMI dataset. The image in the Harmful Memes dataset frequently includes scenes of strikes, fires, group gatherings, and riots. Consequently, the emotional tone, target, and contextual background of these memes diverge significantly from the MAMI dataset, which primarily addresses issues related to targeting and hatred against women. This variation in content underscores the unique challenges posed by the Harmful Memes dataset, affecting the model's ability to generalize across different themes and contexts effectively. Although another prompt-based model, such as PromptHate, is delivering good performance on this dataset, our model, when applied in a zero-shot manner, does not outperform the state-of-the-art (SOTA) models. This highlights areas for further refinement and adaptation of our approach to enhance its performance under zero-shot conditions and across diverse datasets.
\begin{table}[!htp]
\centering
\scriptsize
\begin{tabular}{lrrrr}
\toprule
\textbf{Dataset} & \textbf{Split} & \textbf{Label} & \textbf{\#Memes} \\
\midrule
\multirow{4}{*}{MAMI} & \multirow{2}{*}{Train} & Misogynous &5,000 \\
& & Non-Misogynous &5,000 \\
\cmidrule{3-4}
& \multirow{2}{*}{Test} & Misogynous &500 \\
& & Non-Misogynous &500 \\
\midrule
\multirow{4}{*}{MIMIC} & \multirow{2}{*}{Train} & Misogynous &  2,012\\
& & Non-Misogynous & 2,032 \\
\cmidrule{3-4}
& \multirow{2}{*}{Test} & Misogynous & 503 \\
& & Non-Misogynous & 507\\
\midrule
\multirow{4}{*}{Memotion2} & \multirow{2}{*}{Train} & Offensive &1,933 \\
& & Non-Offensive &5,567 \\
\cmidrule{3-4}
& \multirow{2}{*}{Test} & Offensive &557 \\
& & Non-Offensive &943 \\
\midrule
\multirow{4}{*}{Hateful meme} & \multirow{2}{*}{Train} & Offensive &3,050 \\
& & Non-Offensive &5,450 \\
\cmidrule{3-4}
& \multirow{2}{*}{Test} & Offensive &500 \\
& & Non-Offensive &500 \\
\midrule
\multirow{4}{*}{Harmful meme} & \multirow{2}{*}{Train} & Harmful & 1,064\\
& & Non-harmful & 1,949\\
\cmidrule{3-4}
& \multirow{2}{*}{Test} & Harmful &124  \\
& & Non-Harmful & 230\\
\bottomrule
\end{tabular}
\caption{Class-wise (MAMI, Memotion2, Hateful meme, and HarmMeme dataset) distribution in Train Set and Test Set}
\label{tab:classwise_dis}
\end{table}

\begin{figure*}[t!]
\centering
\includegraphics[width=\linewidth]{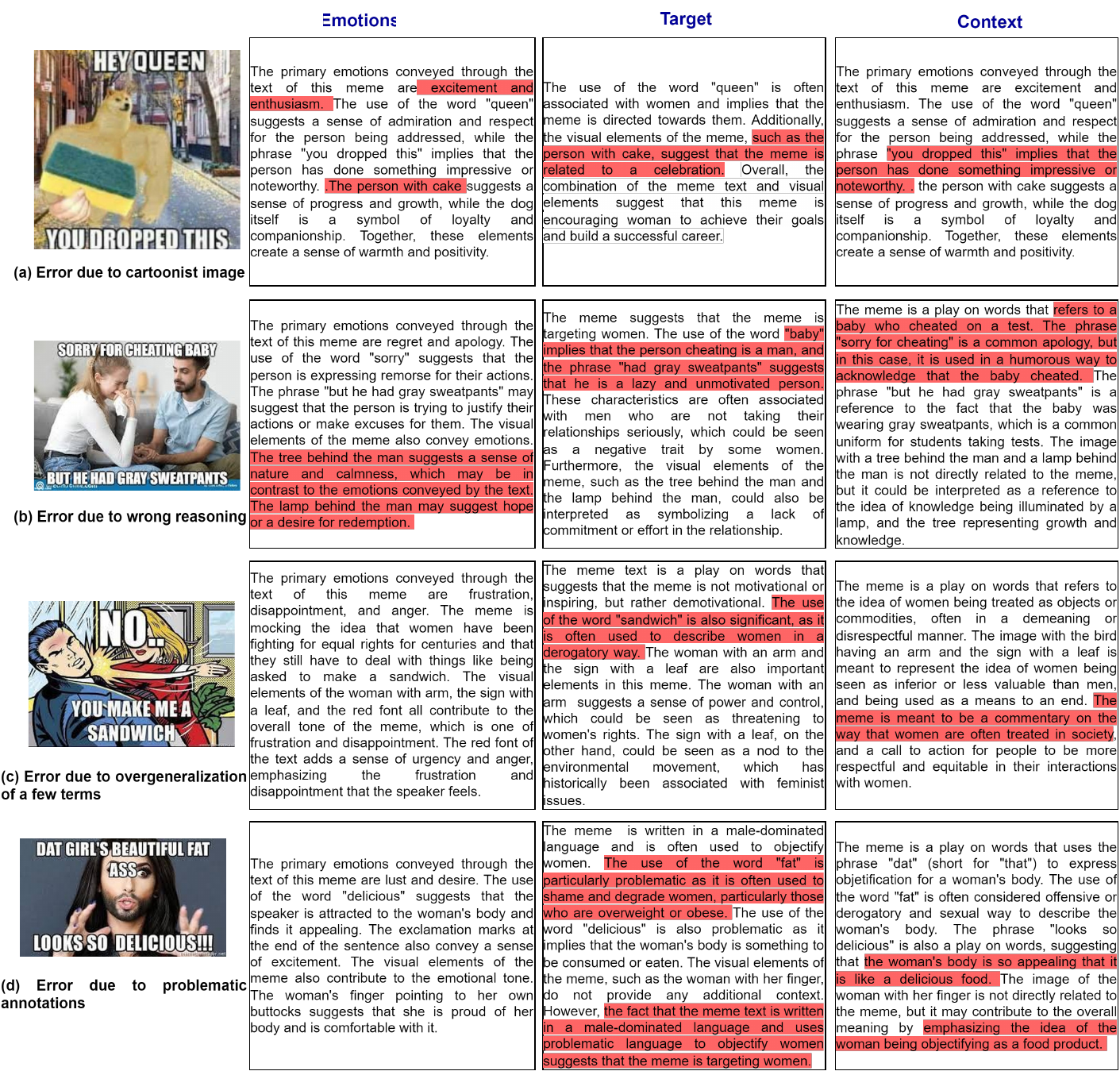}
\caption{ Error analysis of wrong predictions done by our proposed model M3Hop-CoT\label{fig:error_rationale}}
\end{figure*}

\begin{figure*}[t!]
\centering
\includegraphics[width=\linewidth]{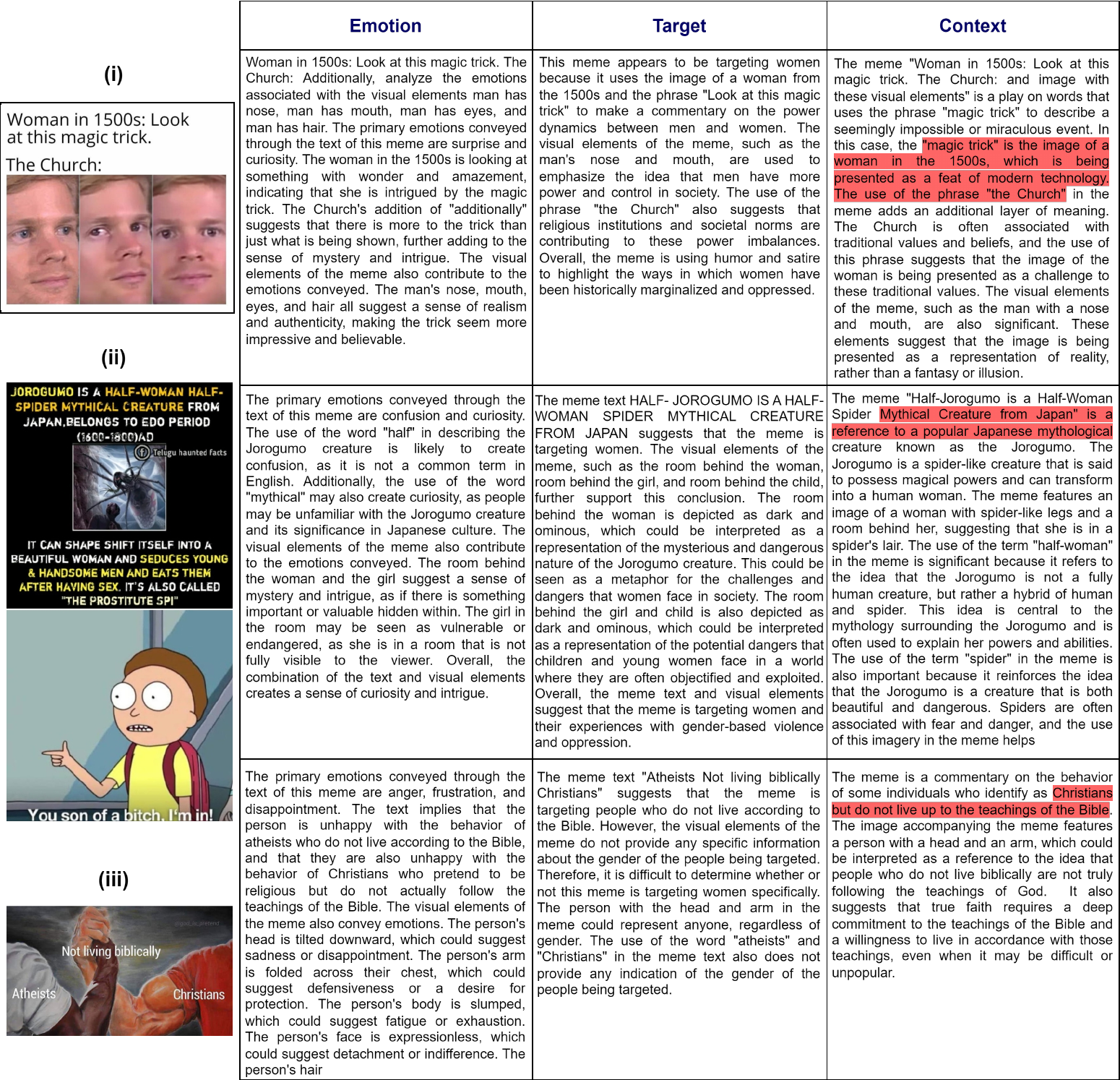}
\caption{ Analysis of rationale generated by the LLMs for the cultural diversity\label{fig:cultural}}
% \vspace{-0.5cm}
\end{figure*}
% \vspace{-0.2cm}
\section{A note on Error Analysis}\label{sec:appen_error}
Among various types of errors, we mostly categorized these errors as (i) Cartoonist image, (ii) Reasoning Failure, (iii) Overgeneralization of identity terms, and (iv) Annotation Error.
As depicted in Appendix Figure \ref{fig:error_rationale} (a), when the scene graph misidentifies objects and their relationships in the meme image, the LLM struggles to accurately correlate EoRs with the meme's text and context, leading to the generation of irrelevant rationales. In this instance, the scene graph mistakenly identifies the scrub sponge for utensil cleaning as a cake, resulting in altering the contextual interpretation. Despite the meme's intention to insult women, the generated rationale paradoxically praises the woman for her achievements. These errors are particularly prevalent in memes featuring cartoonist illustrations or images with unclear visibility.\\ 

\noindent
Beyond scene graph errors, Appendix Figure \ref{fig:error_rationale} (b) showcases another limitation: LLMs can struggle with complex contextual reasoning. Even with accurate EoRs generated by the scene graph, in some scenarios, the LLM fails to understand the overall meaning of the meme. In this example, the meme satirizes the stereotype used to degrade women. However, the LLM misunderstands the context, generating a rationale that targets men for cheating on women. Furthermore, the term ``baby" is used figuratively in the meme, but the LLM interprets it literally, resulting in a misguided rationale. This highlights the ongoing challenge of LLM comprehension when dealing with wordplay, sarcasm, and other forms of nuanced communication within memes.\\

\noindent
The third category of error observed is the overgeneralization of certain keywords identified as identity terms. In Appendix Figure \ref{fig:error_rationale} (c), the LLM misinterprets the context due to this limitation. The presence of the word "sandwich" triggers a bias within the classifier, leading the model to predict the meme as misogynous wrongly. Despite the meme's offensive nature, it does not aim to degrade women; instead, it advocates violence in general. However, the LLM misunderstood the context based on the keyword and generates, \textit{``The use of the word `sandwich' is also significant, as it is often used to describe women in a derogatory way."} showcasing the overgeneralization of this term without contextual consideration.\\ 

\noindent
The last error category we address involves discrepancies arising from subjectivity inherent in the annotation process. Despite the LLM generating accurate rationales aligned with the meme's context, misclassification occurs when the predicted label fails to align with the actual label (Appendix Figure \ref{fig:error_rationale} (d)). Identifying and rectifying such subjectivity is a complex and ongoing area of research. Prior studies on annotation highlight the challenge of effectively mitigating bias and subjectivity despite the implementation of annotation schemes \cite{davidson-etal-2019-racial}. This underscores the need for further exploration and refinement of annotation methodologies to enhance the reliability and objectivity of classification tasks in natural language processing.

\subsection{Further Categorization of Errors }\label{sec:Failure_error}
In Appendix Figure \ref{fig:miss_rate}, we performed a comprehensive error analysis to assess the performance variations of the proposed models across the above-mentioned error categories. CLIP\_MM exhibited limitations in reasoning and comprehending identity terms, demonstrated by a high rate of overgeneralization errors (23.93\%). This suggests potential deficiencies in understanding the diverse identities within the meme text. While CLIP\_MM+GPT4 improved reasoning and identity comprehension, it still struggled with annotation errors, pointing towards potential data labeling issues. Conversely, CLIP\_MM+ChatGPT achieved enhancements across all the metrics, indicating superior overall performance and improved contextual understanding. CLIP\_MM+Llama showed relatively lower overgeneralization and annotation errors, but reasoning failures still exist. Finally, our M3Hop-CoT model achieved the lowest overall error rate, demonstrating significant advancements in reasoning and identity term comprehension. However, annotation errors remain an area for further refinement. These findings highlight the importance of continuous improvement to mitigate errors and enhance the capabilities of these models for tackling complex meme tasks.
 
% \begin{figure}[t!]
% \centering
% \includegraphics[width=\linewidth]{latex/images/Error_rates_CoT.PNG}
% \caption{Misclassification rate comparison between proposed model \textbf{M3Hop-CoT} and their various variants\label{fig:miss_error}}
% \end{figure}
% \begin{figure}[t!]
% \centering
% \includegraphics[width=\linewidth]{ARR_april_error_sample.pdf}
% \caption{ Categorization of error analysis (\%) of proposed model \textbf{M3Hop-CoT} and other SOTA models\label{fig:miss_rate}}
% \end{figure}
\section{Future Works}\label{sec:fututre_work}
While our current zero-shot prompting approach effectively encourages the LLM to generate rationales for misogyny detection, future work could explore fine-tuning the LLM specifically for misogyny detection within memes. Fine-tuning can enhance contextual reasoning by understanding the dataset's pattern, potentially leading to more context-rich rationales and improved misogyny detection. Additionally, it can be particularly beneficial for low-resource domains like misogynous meme identification, offering the potential for superior performance compared to a general-purpose LLM. However, the increased computational cost associated with fine-tuning requires careful consideration, especially when dealing with large datasets or computationally expensive LLM architectures. A future evaluation comparing zero-shot prompting and fine-tuning within our LLM-based model will be crucial for determining the optimal approach for achieving both accurate and efficient misogyny detection in memes.\\
\noindent
Another dimension of this work could be related to the scene graph. In the future, we can aim to improve scene graph analysis to mitigate object and relationship recognition errors within memes. This can done by exploring the creation of dynamic scene graphs that adapt in real time to the evolving themes and symbols within memes.  This can work better for handling cartoon illustrations and low-visibility images.\\
\noindent
Another critical area for future work lies in enhancing the ability of LLMs to perform complex contextual reasoning within the domain of memes. As discussed in the error analysis (c.f. Section \ref{sec:error}), misogyny in memes often relies on subtle cues, wordplay, sarcasm, and other forms of communication.  Current LLMs may struggle to grasp these subtleties, potentially leading to misinterpretations and inaccurate rationale generation. LLMs could benefit from being equipped with pragmatic reasoning techniques that enable them to consider the meme's context, including the speaker's intent, cultural references, and social norms. This would allow the LLM to move beyond the literal meaning of words and understand the underlying message.
\begin{figure*}[h]
    \centering
    \includegraphics[width=0.9\linewidth]{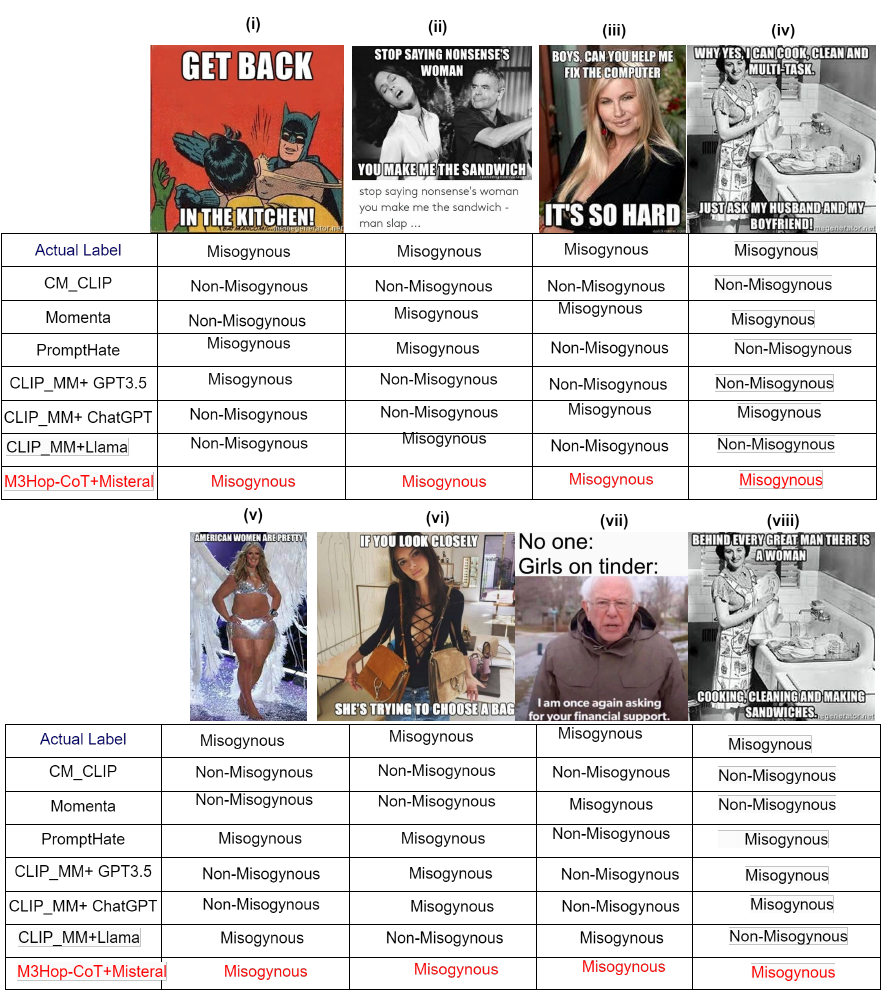}
    % \includepdf[width=\linewidth]{ARR_april_error_llms.pdf}
    % \includegraphics[width=\linewidth]{ARR_april_error_llms.pdf}
    \caption{Predictions from different models for the MAMI Dataset \label{fig:llm_prediction}}
\end{figure*}

\section{Frequently Asked Questions (FAQs)}
\textbf{Que 1: Why did we choose an emotion, target-aware information, and context as the key factors to generate rationale with LLMs.}\\
\textbf{Response:} Extensive previous research demonstrates the critical role of meme emotions in identifying potential toxicity in memes \cite{chauhan-etal-2020-sentiment,Akhtar2022AllinOneES,memotion2,10.1007/978-3-031-28244-7_7,10475492}. These studies also indicate that a major limitation of existing meme identification classifiers is their insufficient contextual understanding \cite{kumari-etal-2021-co}. Our approach leverages Large Language Models (LLMs) to bridge this gap by integrating comprehensive contextual analysis. Furthermore, targeted information is essential for identifying harmful content, as emphasized by \citet{sharma-etal-2022-disarm}. Previous methods primarily relied on supervised learning, which requires extensive data annotation, thereby increasing costs and potential for error. In contrast, our methodology utilizes the capabilities of LLMs to process psycholinguistic features in a cost-effective and error-minimizing manner, thereby enhancing the rationality and effectiveness of meme analysis.\\

\noindent
\textbf{Question 2: Why are the results of the MAMI dataset presented for both the development and test sets?}\\
\textbf{Response:} The MAMI dataset, part of SemEval-2022 Task 5 on Multimedia Automatic Misogyny Identification, aims to explore detecting misogynous memes. This dataset's authors have provided development (dev) and test sets to enable comprehensive evaluation. Presenting results on both sets allows us to assess the model's generalizability and robustness across different subsets of data. This dual-set evaluation strategy ensures that our findings are significant and that the model's performance is robustly demonstrated under varied conditions.\\

\noindent
\textbf{Que 3: Which model serves as the baseline for implementing CoT reasoning?}\\
\textbf{Response:} The baseline model for implementing CoT reasoning is our CLIP-based model, \textit{CLIP\_MM}. This model integrates the CLIP framework's textual and visual encoders to extract respective features from memes. Feature fusion is achieved using the Multimodal Factorized Bilinear (MFB) pooling technique, with a softmax classification layer with two neurons for label prediction. \textit{CLIP\_MM} is optimized using cross-entropy loss and has demonstrated superior performance across all other pre-trained visual-language models for Misogynous meme identification, as evidenced in Table \ref{tab:result}. Its effectiveness establishes it as the foundational model for subsequent enhancements with LLM-based techniques.\\

\noindent
\textbf{Que 4: We have written in the results analysis part: ``We also observed that multimodal baselines give better results than unimodal ones." Isn't it obvious in a scenario like meme analysis?}\\
\textbf{Response:} While it might seem intuitive that multimodal approaches would outperform unimodal ones in meme analysis, this is not universally true. \citet{thomason-etal-2019-shifting} has shown instances where unimodal inputs surpass multimodal ones, often due to noise and interference in multimodal signals, which can obscure rather than clarify the context. The intention behind highlighting this observation in our study was not to restate the obvious but to provide empirical evidence supporting the efficacy of multimodal systems, specifically in meme analysis tasks within our dataset. This empirical validation emphasizes the practicality and effectiveness of using multimodal techniques for handling memes, strengthening our research's findings. But, yes, we agree that multimodal systems should be better than unimodal systems.\\

\noindent
\textbf{Que 5: Why were uniform metrics not employed across all datasets? This approach could potentially enhance the uniformity of the paper.}\\
\textbf{Response:} The four datasets utilized in our evaluation are publicly available and have been widely adopted in SemEval or other competitions hosted by respective organizations, with the exception of the harmful memes dataset. The authors of the original dataset papers employed specific metrics that have since become standard for these datasets. We opted to use the same metrics to facilitate direct comparisons with the state-of-the-art (SOTA) results reported in these papers. This approach ensures that our evaluation is relevant and consistent with existing literature, thereby clearly benchmarking against established results.  \\
\end{document}